\documentclass[letterpaper]{article}

\PassOptionsToPackage{numbers, compress}{natbib}

\usepackage[preprint]{neurips_2026}

\usepackage[utf8]{inputenc} 
\usepackage[T1]{fontenc}    
\usepackage{hyperref}       
\usepackage{url}            
\usepackage{booktabs}       
\usepackage{amsfonts}       
\usepackage{nicefrac}       
\usepackage{microtype}      
\usepackage{xcolor}         
\usepackage{kotex}

\usepackage[table]{xcolor}
\usepackage[most]{tcolorbox}
\usepackage{array}

\usepackage{makecell}
\usepackage{hhline}
\usepackage{bm}
\usepackage{subcaption}

\definecolor{lumosblue}{RGB}{0,0,0}

\newcolumntype{C}[1]{>{\centering\arraybackslash}p{#1}}

\usepackage{svg}
\usepackage{kotex}
\usepackage{wrapfig}
\usepackage{graphicx}
\usepackage{multirow}
\usepackage{arydshln}
\usepackage{enumitem}
\usepackage{subcaption}
\usepackage[textsize=tiny]{todonotes}
\usepackage[capitalize,noabbrev]{cleveref}

\usepackage{amsmath}
\usepackage{amssymb}
\usepackage{mathtools}
\usepackage{amsthm}

\theoremstyle{plain}

\theoremstyle{definition}

\theoremstyle{remark}

\newcommand{\method}{\textsc{Lumos}}

\title{\method: Tracing Parametric Knowledge from Training Data to Behavioral Outputs in LLMs}

\author{%
  \begin{tabular}{@{}c@{\hspace{2.5em}}c@{\hspace{2.5em}}c@{}}
    \textbf{Seoyeon Ye}\textsuperscript{1}\thanks{Equal contribution} &
    \textbf{Gayoung Kim}\textsuperscript{1}\footnotemark[1] &
    \textbf{Jiyoung Hong}\textsuperscript{1}\footnotemark[1]
  \end{tabular}
  \\[0.6em]
  \begin{tabular}{@{}c@{\hspace{3em}}c@{}}
    \textbf{Sookyung Kim}\textsuperscript{1,2}\thanks{Corresponding authors} &
    \textbf{Hyunsoo Cho}\textsuperscript{1,2}\footnotemark[2]
  \end{tabular}
  \\[0.7em]
  \textsuperscript{1}Division of Artificial Intelligence \& Software,
  Ewha Womans University \\
  \textsuperscript{2}Institute for Multiscale Matter and Systems,
  Ewha Womans University \\
  \texttt{\{s25y25n,chohyunsoo\}@ewha.ac.kr}
}

\begin{document}

\maketitle

\begin{abstract}


Current analyses of LLMs' parametric knowledge are largely output-centric, drawing conclusions about what a model \textit{knows} without verifying what it was actually trained on.
This leaves fundamental questions, such as whether a correct response reflects genuine generalization or rote memorization, grounded in speculation rather than evidence.
To resolve these ambiguities, we introduce \method \footnote{The code is available at: \url{https://github.com/scribbleNdoodle/lumos}}, a diagnostic 
framework that traces knowledge
\textcolor{lumosblue}{across training-data exposure, internal representations, and behavioral outputs,}
leveraging OLMo 2 with its fully transparent training corpus.
By grounding analysis in verified exposure, we reveal that models internally encode rare facts with high separability (84\%) yet fail to express them behaviorally (54\%), though this retrieval gap narrows with scale.
Furthermore, when models are asked to self-reflect on their own answers, they perform reliably on trained content (83\%) but drop to random-baseline levels (49\%) on unseen content. 
This collapse persists even under chain-of-thought prompting, which inflates confidence signals rather than improving calibration.
Collectively, these findings demonstrate that incorporating the training-data axis into LLM evaluation transforms speculative diagnoses into 
verifiable claims, and we advocate that this axis should be a standard component of knowledge assessment in LLMs.

\end{abstract}
\section{Introduction}
\label{sec:intro_en}

    LLMs are now the world’s largest repositories of parametric knowledge.
    This knowledge, unlike traditional databases, is not \textit{explicitly retrieved} but is \textit{implicitly reconstructed} through complex neural interactions during inference~\citep{petroni2019language}.
    This implicit and high-dimensional encoding, coupled with the stochastic nature of decoding, creates a fundamental `interpretability gap':
    while we can observe a model's fluent output, we remain \textit{blind} to whether that output stems from a robust internal understanding or a lucky stochastic coincidence \citep{berglund2023reversal, zhangcomprehension, jacovi2020towards}.
    Consequently, the inability to map a model's output back to its training origins represents a fundamental challenge to reliably assessing what LLMs actually know~\citep{ngoalignment, grosse2023studying, wu2024enhancing}.

    To bridge this gap, the research community has developed several probing techniques:
    the most common approach, \textit{behavioral probing}, assesses knowledge through observable outputs ($y$) such as benchmark accuracy \cite{trienes-etal-2025-behavioral, hendrycksmeasuring} or logit-level probability distributions \cite{kadavath2022language, jiang2021can}. 
    To look beyond outputs, \textit{internal probing} examines hidden representations ($h$) and probabilistic preferences ($p$) using diagnostic classifiers \cite{belinkov2022probing, burnsdiscovering, tenney2019bert, alain2016understanding}. 
    More recently, \textit{mechanistic probing} utilizes techniques such as activation patching \cite{vig2020investigating, vig2020causal, zhang2024carefulexaminationlargelanguage} or model editing \cite{hernandez2023linearity, dai2022knowledge, meng2022locating, geva2021transformer, bau2017network} to isolate the specific parameters responsible for factual recall.
    Together, these approaches probe different stages of knowledge manifestation, from internal encoding ($h, p$) to final output ($y$), yet they all omit a critical preceding stage: the training data ($D$) that shaped these representations in the first place.
    Without this grounding, we cannot distinguish memorization from generalization, separate knowledge absence from retrieval failure, or tell whether model confidence reflects genuine understanding or frequency artifacts.

\begin{figure*}[t]
    \centering
    \includegraphics[width=0.99\textwidth]{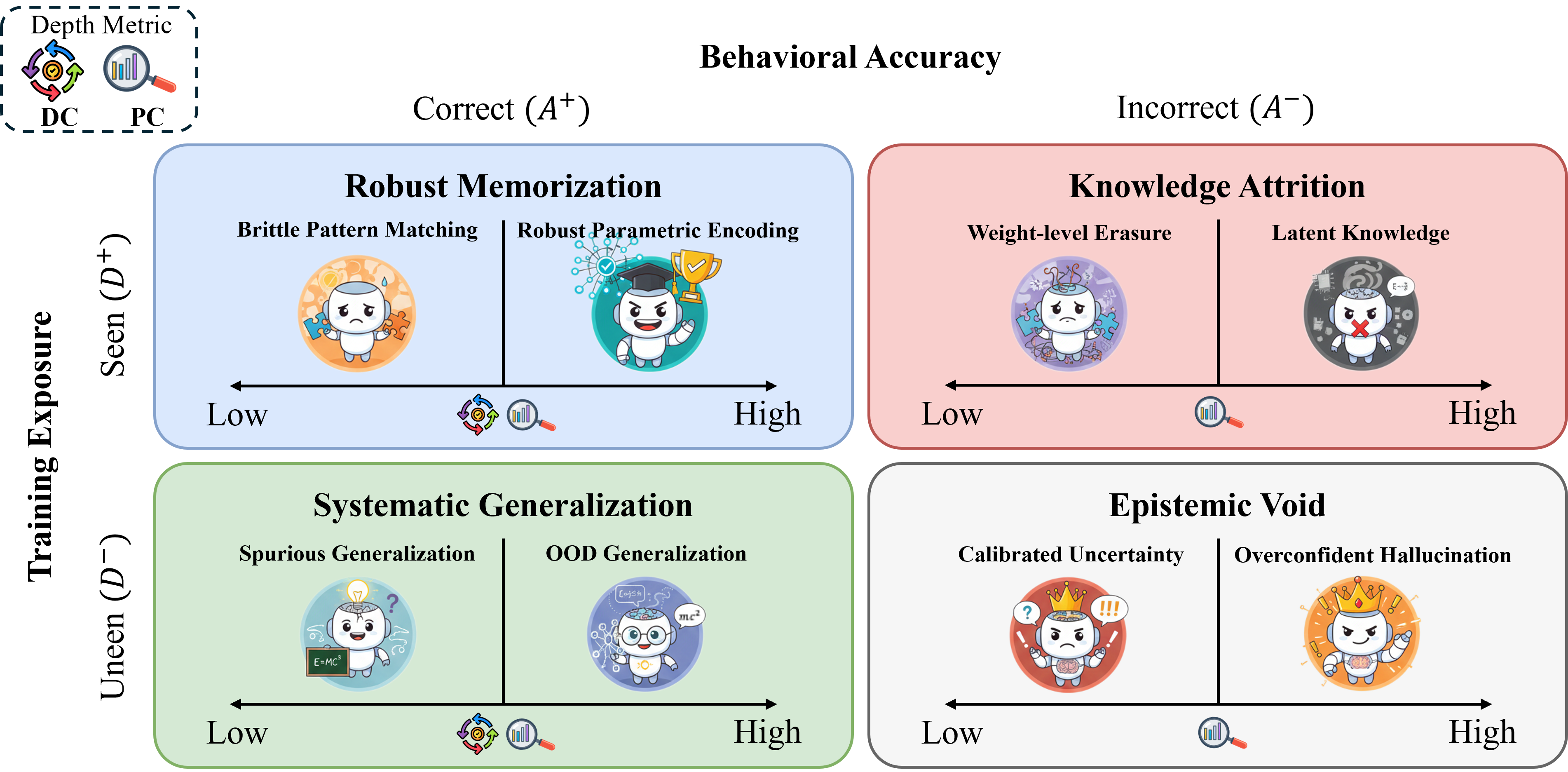}
    \caption{A 2×2 taxonomy of model knowledge states defined by data exposure ($\mathcal{D}$) and behavioral accuracy ($A$), enriched with parametric and decoding depth (Probabilistic Confidence ($PC$), Decisional Consistency ($DC$)).}
    \label{fig:example}
\end{figure*}

    In this work, we introduce \method, a diagnostic framework that completes this chain by grounding evaluation in what the model was actually trained on.
    As illustrated in Figure~\ref{fig:example}, \method~traces knowledge along the full path $D \to h \to p \to y$ and organizes analysis along two axes: \textbf{Training Exposure} (Seen vs. Unseen) and \textbf{Behavioral Accuracy} (Correct vs. Incorrect).
    We instantiate this framework on OLMo 2 (7B--32B) with its fully transparent training corpus, constructing paired seen/unseen benchmarks across factual and reasoning domains.
    To ensure a fine-grained diagnosis, we stratify factual knowledge by training-time frequency (head vs. tail entities), vary prompting strategies (zero-shot, few-shot, and chain-of-thought), and evaluate models through both behavioral metrics and multiple internal probing methods and activation-based prediction.
    For reasoning, we compare RLVR and SFT variants across mathematical and code domains to isolate the effect of training methodology on generalization.
    This multi-dimensional analysis reveals the following findings:
    
    First, models internally encode rare facts with high Linear Probing scores yet fail to express them in their outputs, revealing a retrieval gap in the $h \to y$ transition.
    Second, when models are asked to judge whether their own answers are correct, they perform reliably on trained content but collapse to random-baseline levels on unseen content. 
    Chain-of-thought prompting does not resolve this failure; instead, it inflates confidence signals without improving calibration.
    Third, factual reliability is fundamentally anchored in training exposure rather than model scale: even at 32B, accuracy on unseen content remains below 50\%, while seen content consistently exceeds 80\%. Few-shot prompting narrows this gap only for already-encoded knowledge, confirming that the exposure boundary represents a structural limit that neither scaling nor prompting strategies can overcome.
    Finally, an exposure-grounded comparison of RLVR and SFT reveals that their generalization behaviors differ qualitatively across scales: RLVR builds genuine reasoning capability at smaller scales, while at larger scales its advantage reduces to sampling efficiency. Moreover, this pattern is domain-dependent, as RLVR's gains diminish in code generation where reward-task alignment is weaker.

    To summarize, we introduce \method, a framework and benchmark suite that grounds LLM knowledge evaluation in training exposure, turning previously conjectured phenomena into empirically verifiable claims and uncovering failure modes that output-centric evaluation alone cannot detect.

\section{Related Work}
\label{sec:02related_work_new}


\textbf{Behavior Probing (LLM output $y$)} assesses knowledge through observable outputs, such as accuracy \citep{trienes-etal-2025-behavioral, hendrycksmeasuring} or confidence estimates \citep{han2025mind, mahaut2024factual}. Since discrete generation is often stochastic and prompt-sensitive, recent studies seek more stable signals by aggregating behaviors across paraphrased prompts \citep{zhao2024belief}, controlled knowledge conditions \citep{kim2025uniknow}, or many-shot sampling \citep{wu2024latent}. These methods mainly characterize model uncertainty and selective answering behavior using predictive entropy \citep{quevedo2024detectinghallucinationslargelanguage}.

\textbf{Internal Probing (hidden vector $h$ or logits $p$)} examines knowledge by accessing model internals, primarily through output distribution analysis or latent representation probing. The former uses token-level logits to measure calibration and epistemic uncertainty, showing that probabilities assigned to correct answers (e.g., $P(\text{True})$) correlate more reliably with factual accuracy than surface-form text \citep{kadavath2022language, lin2022teachingmodelsexpressuncertainty}. To disentangle uncertainty from linguistic variability, later work focuses on semantic entropy across answer clusters \citep{kuhn2023semanticuncertaintylinguisticinvariances}. Alternatively, researchers investigate latent representations via diagnostic classifiers trained on hidden states \citep{belinkov2022probing, azaria2023internal}. Such studies consistently reveal an ``expression gap,'' where models encode internal information that fails to manifest in final outputs \citep{gekhman2025inside, orgad2024llms}.


Despite their differences, these lines of work share a critical blind spot: they rarely ground their conclusions in what the model \emph{actually saw} during training.

\section{\method\ Taxonomy}
\label{sec:03methods}

\textcolor{lumosblue}{We organize the analysis of parametric knowledge using a conceptual sequence of stages:}

\vspace{-0.8em}
\begin{equation*}
\mathcal{D} \xrightarrow{\text{exposure}} \mathbf{h} \xrightarrow{\text{probing}} p \xrightarrow{\text{decoding}} y
\label{eq:chain}
\end{equation*}
\vspace{-0.8em}

where $\mathcal{D}$ denotes the training corpus (and its measurable exposure statistics), $\mathbf{h}$ the model's internal representations (hidden states), $p$ the resulting probabilistic preference (logits), and $y$ the final behavioral output.
\textcolor{lumosblue}{
This sequence serves as a conceptual framework for organizing our analysis rather than an empirically validated causal pathway.
}


\textcolor{lumosblue}{
Our central principle is \textbf{training-data alignment}:
we characterize knowledge-related behavior by jointly examining (i) evidence of exposure in $\mathcal{D}$, (ii) the model's ability to reliably express information in $y$, and (iii) its internal uncertainty and decodability signals.
}

\subsection{Dual-Axis: Exposure vs. Performance}
To systematically analyze the origins of model behavior, we introduce 
two independent axes: \textbf{Training Exposure ($\mathcal{D}$)} and 
\textbf{Behavioral Accuracy ($A$)}, where $A(y) \in \{A^+, A^-\}$ 
denotes binary correctness against the ground-truth. As illustrated in Figure~\ref{fig:example}, their intersection 
yields four quadrants:

\textbf{Type I: Robust Memorization ($\mathcal{D}^+, A^+$):} 
Facts present in $\mathcal{D}$ and correctly recalled.

\textbf{Type II: Knowledge Attrition ($\mathcal{D}^+, A^-$):} 
Facts present in $\mathcal{D}$ but failing in $y$.

\textbf{Type III: Systematic Generalization ($\mathcal{D}^-, A^+$):} 
Correct responses for facts absent from $\mathcal{D}$.

\textbf{Type IV: Epistemic Void ($\mathcal{D}^-, A^-$):} 
Expected failures defining the model's true boundaries.

\subsection{\textcolor{lumosblue}{Depth Axis: Operationalizing Epistemic Depth}}

While the dual-axis framework categorizes knowledge states, a binary $A(y)$ is insufficient to distinguish between robust internal understanding or lucky coincidence.
Since LLM decoding is inherently non-deterministic, the same query can yield various outputs, even when the underlying parametric knowledge is unchanged.
\textcolor{lumosblue}{
To quantify the continuous spectrum ranging from robust, low-variance knowledge to fragile, high-variance behavior within each quadrant, we introduce a third axis—Epistemic Depth—which analyzes behavioral consistency and diagnostic signals associated with internal states ($\mathbf{h}, p$) and the final output $y$.
}
Specifically, for each query $q$, we execute $N$ stochastic forward passes to extract the corresponding internal states and behavioral outputs:
\begin{equation} {(\mathbf{h}^{(i)}, p^{(i)}, y^{(i)})}_{i=1}^{N}, \quad \text{where } y^{(i)} \sim \pi(\cdot \mid p^{(i)}) 
\end{equation}
Here, each $y^{(i)}$ is sampled from the probabilistic preference $p^{(i)}$ induced by the latent representation $\mathbf{h}^{(i)}$. 
\textcolor{lumosblue}{This triplet allows us to examine the alignment and variability across internal states, probabilistic preferences, and behavioral outputs.}

\paragraph{Decisional Consistency ($DC$) Probing $p \to y$:}
$DC$ evaluates the robustness of the final decoding stage by measuring how consistently the model's probabilistic preferences $p^{(i)}$ converge to the same behavioral outcome:
\begin{equation}
DC(q) \;=\; \frac{1}{N} \sum_{i=1}^{N} \max_{y \in \mathcal{Y}}  n(y),
\end{equation}
where $n(y)$ is a normalized frequency of the option $y\in\mathcal{Y}$ for question $q$.
Intuitively, a high $DC$ indicates that the model consistently produces the same output across multiple samples, regardless of its correctness, reflecting a stable mapping from preferences to final decisions.

\paragraph{Probabilistic Confidence ($PC$) Probing $p$, $\mathbf{h} \to p$:}
$PC$ quantifies \textcolor{lumosblue}{indicators of} epistemic grounding of internal states. We define an uncertainty operator $\mathcal{U}$ that analyzes the quality of the mapping from latent activations ($\mathbf{h}$) to probabilistic preferences ($p$):
\begin{equation}
PC(q) = \mathcal{U}\big(p, \mathbf{h}\big)
\end{equation}
Here, $\mathcal{U}$ can be instantiated as either Predictive Uncertainty (e.g., logit entropy) or Latent Uncertainty (e.g., diagnostic probing scores of $\mathbf{h}$).
\textcolor{lumosblue}{This allows us to assess whether internal signals are consistent with well-grounded, high-confidence representations or suggest a more fragile state despite achieving behavioral success.}
In particular, we employ four complementary quantification metrics (introduced in Section~\ref{sec:eval-metrics}) to operationalize $PC(q)$.
We additionally collect \emph{self-reported} confidence via prompting, and use it as auxiliary behavioral metadata.

\subsection{Multidimensional Knowledge Diagnostics}
We synthesize the four-quadrant framework with the epistemic depth axis ($PC, DC$) to define the model's internal knowledge states. This multidimensional mapping allows us to distinguish between grounded parametric knowledge and superficial behavioral patterns.

\textbf{Type I (Robust Memorization, 
$\mathcal{D}^+, A^+$):} 
High $PC$ and $DC$ indicate \textit{Robust Parametric Encoding}, where knowledge is deeply embedded and stable. Low depth metrics instead reveal \textit{Brittle Pattern Matching} via linguistic shortcuts.

\textbf{Type II (Knowledge Attrition, $\mathcal{D}^+, A^-$):}
High $PC$ identifies \textit{Latent Knowledge}, where the parametric representation is preserved in the hidden states but fails to manifest during the $p \to y$ decoding transition. Low $PC$ confirms \textit{Weight-level Erasure}, signifying that the information has been fundamentally purged from the model’s parameters during the optimization process.

\textbf{Type III (Systematic Generalization, $\mathcal{D}^-, A^+$):}
High depth metrics validate \textit{OOD Generalization}, confirming that the model has synthesized invariant logical rules from unseen data. Low metrics indicate \textit{Spurious Generalization}, where correct outputs arise from accidental alignment rather than grounded reasoning.

\textbf{Type IV (Epistemic Void, $\mathcal{D}^-, A^-$):}
Low $PC$ reflects \textit{Calibrated Uncertainty}, where the model's internal signals accurately reflect its lack of knowledge. High $PC$, by contrast, flags \textit{Overconfident Hallucination}—a critical failure mode where false outputs are generated with unwarranted internal confidence.

Our empirical analysis focuses particularly on Type II and Type IV, where the interplay between training exposure and epistemic depth reveals actionable failure modes such as retrieval bottlenecks and miscalibrated self-assessment.

\section{\method\ Dataset}\label{sec:experiment-design}
Our analysis requires full access to both model internals and the pretraining corpus.
We therefore use the \emph{OLMo~2-Instruct} family, whose fully transparent training data enables precise verification of training exposure and exposure-grounded analysis of model behavior.

\subsection{Evaluation Domains: Factual vs. Reasoning}
\label{sec:evaluation-domains}

To conduct a granular diagnostic across our four epistemic quadrants (Type I--IV), we partition our evaluation into two primary domains:
\textbf{Factual} and \textbf{Reasoning}. 
This categorical split is essential for isolating distinct learning mechanisms in LLMs. 
The Factual domain, containing knowledge such as ``\textit{The capital of the U.S. is Washington D.C.,}'' serves to measure the stability of parametric memory. 
In this domain, success on unseen data is often attributable to accidental exposure or spurious guessing rather than logical deduction.
In contrast, the Reasoning domain—encompassing mathematics, logic and programmatic reasoning—allows for a meaningful assessment of systematic generalization. To further analyze model behavior within this setting, we additionally include the OLMo 2-SFT family alongside OLMo 2-Instruct for comparative evaluation. This setup enables us to isolate how reasoning behavior differs across models that share the same pretraining corpus but adopt different post-training paradigms, namely supervised fine-tuning versus reinforcement-based alignment.
By reporting results independently for each domain, we aim to provide a more rigorous and nuanced analysis of model behavior.
To operationalize our framework, we construct a paired QA dataset consisting of Seen and Unseen QA pairs from the OLMo~2 pretraining corpus (summarized in Table~\ref{tab:datasets_overview} Seen split).

\begin{table}[t]
\centering
\caption{Datasets used for factual and reasoning QA evaluation.
Seen datasets are drawn from corpora explicitly used during training,
while unseen datasets are excluded from training.}
\label{tab:datasets_overview}

\resizebox{0.95\textwidth}{!}{


\small
\renewcommand{\arraystretch}{1.1}
\begin{subtable}[t]{0.50\linewidth}
\centering
\caption{\textbf{Seen}}
\resizebox{\linewidth}{!}{%
\setlength{\tabcolsep}{11pt}
\begin{tabular}{@{}lllr@{}}
\toprule
\textbf{Knowledge} & \textbf{Domain} & \textbf{Dataset} & \textbf{\# Sample} \\
\midrule
\multirow{3}{*}{Factual}
  & \multirow{2}{*}{Web} & DCLM       & \multirow{2}{*}{1,387} \\
  &                      & Wikipedia  &                         \\
\cmidrule(lr){2-4}
  & Academic             & peS2o      & 366 \\
\midrule
Reasoning & Math & GSM8K (Train) & 1,319 \\
          &      & MetaMath-Owmfilter & 1,000 \\
\bottomrule
\end{tabular}
}
\end{subtable}
\hfill
\begin{subtable}[t]{0.44\linewidth}
\centering
\caption{\textbf{Unseen}}
\resizebox{\linewidth}{!}{%
\begin{tabular}{@{}lllr@{}}
\toprule
\textbf{Knowledge} & \textbf{Domain} & \textbf{Dataset} & \textbf{\# Sample} \\
\midrule
\multirow{2}{*}{Factual} & \multirow{2}{*}{Web} & RealTime QA    & 220 \\
        &     & Current events & 524 \\
\midrule
\multirow{3}{*}{Reasoning}
  & \multirow{3}{*}{Math} & GSM8K (Test) & 1,319 \\
  &                       & GSM1K        & 1,205 \\
  &                       & GSM-Symbolic & 1,000 \\
\bottomrule
\end{tabular}
}
\end{subtable}
}
\end{table}

\subsection{Factual Domain Construction}
\label{sec:factual-construction}
\textbf{Seen Dataset: Web and Academic Subsets} We further stratify the factual domain into Web and Academic subsets, as they exhibit different structural properties.
For the Web domain, we use a 10\% stratified sample chunk of the \emph{DCLM}~\citep{li2024datacomp}  and the Wikipedia subset of \emph{Dolma 1.7.}~\citep{soldaini2024dolma} For the Academic domain, we utilize the \emph{peS2o}~\citep{soldaini2023pes2o} dataset. 
From the sampled chunks, we extract entities and their associated frequencies through the following pipeline.
We organize queries around entity-centric questions, using \emph{DBpedia}~\citep{lehmann2015dbpedia} for the Web domain and \emph{OpenAlex} Concepts~\citep{priem2022openalex} for Academic.
We employ an Aho--Corasick automaton~\citep{aho1975efficient} to perform exact-string matching across the entire pretraining mix to calculate precise exposure counts.
To analyze how frequency affects knowledge stability, we rank entities by their evidence volume.
Following the definition in~\citep{zong2025comparisonqaevaluatingfactualityrobustness}, high-frequency (\textbf{Head}) entities are defined as those whose cumulative relationships account for the top one-third of all entities when sorted by frequency. Low-frequency (\textbf{Tail}) entities are then sampled from those appearing exactly once in the entire pretraining corpus, matched in size to the Head group. This grouping enables us to contrast model behavior across distinct regions of the frequency distribution
(The full histogram is shown in Appendix Fig.~\ref{fig:entity-frequency}).
For each selected segment, we generate an entity-grounded Multiple-Choice Question (MCQ) using Qwen 2.5 32B following the CRAFT framework~\cite{hossain2025craft}. We apply a Cross-Encoder (\texttt{ms-marco-MiniLM-L12-v2}) to filter for high-density, factually substantive chunks.

\textbf{Unseen Dataset: Temporal Sources}
To ensure a true ``Epistemic Void,'' we curate queries from two sources: ~\citep{kasai2023realtime} and Wikipedia articles covering recent and ongoing current events. All queries are restricted to events occurring strictly after the OLMo~2 release cutoff (March 25, 2025).

\textcolor{lumosblue}{We distinguish between a cutoff-only unseen set, defined by this temporal criterion, and a failure-filtered subset designed to reduce potential contamination from pre-cutoff knowledge.}
For the latter, each multiple-choice query is reformulated as an open-ended question and evaluated using \textcolor{lumosblue}{OLMo~2-Instruct 7B, 13B, and 32B, with $N=11$ generations per model. We exclude any query answered correctly at least once by any of the three models, retaining only instances consistently answered incorrectly across all models and generations.}
The exact prompting template used for this filtering procedure is provided in Appendix~\ref{prompts:filtering}.

\textcolor{lumosblue}{Although this procedure aims to remove potential contamination from pre-cutoff knowledge, it also introduces a selection bias toward questions that the evaluated models fail to answer. Consequently, the filtered set may not be representative of all temporally unseen questions. We use this filtered set for our main analysis to minimize potential contamination and report results on the cutoff-only unseen set in the Appendix~\ref{sec:cutoff_only} for comparison.}

\begin{table*}[t]
  \centering
  \caption{Factual evaluation across Seen/Unseen splits under standard prompting (zero-shot and few-shot). Self-verify denotes the accuracy computed based on the corresponding confusion matrix. Numbers in parentheses denote relative ranks (lower is better). \textbf{Boldface} highlights values used for within-table comparisons, whereas \underline{underlining} indicates values compared across tables.} 
  \label{tab:RQ1_factual}

  \begingroup
  \setlength{\tabcolsep}{3pt}
  \renewcommand{\arraystretch}{1.15}
  \small

  \resizebox{\textwidth}{!}{%
    \setlength{\tabcolsep}{6pt}
\renewcommand{\arraystretch}{1.15}
\resizebox{\textwidth}{!}{
\begin{tabular}{
  >{\centering\arraybackslash}m{0.75cm}
  >{\centering\arraybackslash}m{1.35cm}
  l
  c c >{\columncolor{gray!8}}c >{\columncolor{gray!8}}c
  c c >{\columncolor{gray!8}}c >{\columncolor{gray!8}}c
}

\toprule
& & & \multicolumn{4}{c}{\textbf{Zero-shot}}
    & \multicolumn{4}{c}{\textbf{Few-shot (5)}} \\
\cmidrule(lr){4-7} \cmidrule(lr){8-11}

& & &
\multicolumn{2}{c}{Seen} &
\multicolumn{2}{c}{Unseen} &
\multicolumn{2}{c}{Seen} &
\multicolumn{2}{c}{Unseen} \\
\cmidrule(lr){4-5}
\cmidrule(lr){6-7}
\cmidrule(lr){8-9}
\cmidrule(lr){10-11}

\textbf{\textbf{OLMo~2}} & \textbf{Types} & \textbf{Metrics}
& \makecell{\textbf{Academic}}
& \makecell{\textbf{Web}}
& \makecell{\textbf{RealTime}\\\textbf{QA}}
& \makecell{\textbf{Current}\\\textbf{events}}
& \makecell{\textbf{Academic}}
& \makecell{\textbf{Web}}
& \makecell{\textbf{RealTime}\\\textbf{QA}}
& \makecell{\textbf{Current}\\\textbf{events}} \\
\midrule

\multirow{8}{*}{\makecell{\textbf{7B}}}
& \multirow{3}{*}{\makecell{\textbf{Behavioral}}}
& Accuracy        & 75.53 (17) & \textbf{70.09 (23)} & \underline{\textbf{33.88 (45)}} & 32.92 (46) & 78.99 (11) & \textbf{76.04 (14)} & \textbf{35.70 (41)} & 36.62 (39) \\
&  & $DC$     & 86.75 (17) & 83.36 (22) & 74.49 (36) & 73.01 (41) & 91.47 (5) & 88.96 (14) & 79.28 (30) & 78.78 (31) \\
&  & Self-Verify  & 28.00 (46) & 34.29 (42) & 61.27 (15) & 66.30 (10) & 26.52 (47) & 29.77 (45) & 59.94 (19) & 63.61 (11) \\
\cmidrule{2-3}
& \multirow{4}{*}{\makecell{\textbf{Internal}\\($PC$)}}
& Norm. $P$ & 63.19 (15) & 57.49 (19) & \underline{28.18 (41)} & 24.81 (43) & 63.44 (14) & 56.82 (21) & 27.27 (42) & 23.42 (44) \\
&  & MARS            & 58.73 (24) & 60.00 (18) & 51.61 (45) & 50.34 (46) & 58.45 (27) & 62.20 (13) & 52.73 (43) & 58.46 (26) \\
&  & Linear Probing  & 80.23 (10) & \textbf{83.31 (5)} & 60.94 (31) & 64.34 (29) & 75.42 (18) & 81.50 (7) & 54.74 (40) & 59.80 (32) \\
&  & SAPLMA          & 75.72 (6) & 83.24 (3) & 34.55 (11) & 52.93 (8) & - & - & - & - \\
\addlinespace[2pt]
\midrule

\multirow{8}{*}{\makecell{\textbf{13B}}}
& \multirow{3}{*}{\makecell{\textbf{Behavioral}}}
& Accuracy        & 79.78 (7) & \textbf{74.33 (19)} & 38.18 (36) & 36.36 (40) & 79.02 (10) & 75.06 (18) & 36.64 (38) & 35.14 (43) \\
&  & $DC$     & 91.07 (7) & 85.19 (21) & 71.85 (44) & 71.11 (45) & 89.89 (10) & 86.36 (19) & 72.95 (42) & 72.22 (43) \\
&  & Self-Verify  & 62.54 (12) & 55.70 (21) & 49.53 (32) & 48.42 (36) & 60.89 (17) & 57.18 (20) & 49.53 (32) & 48.02 (37) \\
\cmidrule{2-3}
& \multirow{4}{*}{\makecell{\textbf{Internal}\\($PC$)}}
& Norm. $P$ & 29.69 (37) & 29.54 (39) & 15.72 (47) & 17.04 (45) & 29.60 (38) & 29.36 (40) & 14.48 (48) & 16.55 (46) \\
&  & MARS            & 64.84 (2) & 63.71 (4) & 49.99 (47) & 53.32 (41) & 64.17 (3) & 65.97 (1) & 58.17 (28) & 60.30 (16) \\
&  & Linear Probing  & 76.15 (17) & \textbf{82.53 (6)} & 50.75 (46) & 59.51 (33) & 80.89 (8) & 83.55 (4) & 49.13 (48) & 55.13 (39) \\
&  & SAPLMA          & 82.97 (4) & 84.40 (2) & 27.88 (12) & 48.60 (9) & - & - & - & - \\
\addlinespace[2pt]
\midrule

\multirow{8}{*}{\makecell{\textbf{32B}}}
& \multirow{3}{*}{\makecell{\textbf{Behavioral}}}
& Accuracy        & \textbf{82.97 (1)} & \textbf{80.22 (5)} & \textbf{43.64 (29)} & 43.81 (28) & 82.41 (2) & 79.92 (6) & \textbf{43.20 (31)} & 42.54 (32) \\
&  & $DC$     & \textbf{94.73 (1)} & 89.72 (11) & \textbf{78.73 (32)} & 79.81 (29) & 92.56 (4) & 87.90 (15) & 75.15 (34) & 73.75 (38) \\
&  & Self-Verify  & 83.76 (2) & 79.67 (7) & 54.77 (23) & 53.48 (26) & 83.83 (1) & 80.53 (5) & 53.88 (24) & 53.57 (25) \\
\cmidrule{2-3}
& \multirow{4}{*}{\makecell{\textbf{Internal}\\($PC$)}}
& Norm. $P$ & 57.06 (20) & 54.15 (24) & 30.17 (36) & 31.78 (34) & 57.52 (18) & 54.73 (23) & 30.54 (35) & 33.72 (33) \\
&  & MARS            & 56.91 (33) & 62.45 (12) & 48.09 (48) & 52.34 (44) & 56.00 (38) & 62.86 (7) & 57.35 (30) & 60.06 (17) \\
&  & Linear Probing  & \textbf{71.32 (23)} & \textbf{78.29 (15)} & \textbf{50.55} (47) & 51.10 (44) & 75.41 (19) & 85.31 (2) & 52.78 (43) & 70.25 (25) \\
&  & SAPLMA          & 82.61 (5) & 86.19 (1) & 40.61 (10) & 59.54 (7) & - & - & - & - \\
\bottomrule

\end{tabular}}

  }
  
  \endgroup
\end{table*}

\subsection{Reasoning Domain Construction}
\label{sec:reasoning-construction}

To cover a broad spectrum of reasoning, we structure our evaluation across two domains: mathematical and code-based reasoning. We focus on mathematical reasoning in this section; code-based reasoning, where reward-task alignment differs qualitatively, serves a complementary role and is presented in Appendix~\ref{sec:Results_in_Code_Dataset}.

\textbf{Seen Dataset} We use the \textit{GSM8K}~\citep{cobbe2021training} training split, which was explicitly included in the OLMo~2 training mix. We also include \textit{MetaMath-Owmfilter}, a filtered \emph{MetaMathQA} subset from OLMo~2's \emph{Dolmino-Mix-1124} math mixture~\citep{yu2024metamath,ai2_2025_olmo2}. These datasets evaluate the reasoning patterns and numerical templates the model was directly exposed to during optimization.

\textbf{Unseen Dataset} To measure genuine generalization (Type III), we utilize the \textit{GSM8K} test split along with \textit{GSM1K}~\citep{zhang2024careful} and \textit{GSM-Symbolic}~\citep{mirzadeh2025gsmsymbolicunderstandinglimitationsmathematical}. These datasets are structurally aligned with the training templates but employ modified numerical values and entities. This ensures that success is contingent upon the internalization of logical rules rather than the memorization of specific answer tokens.

\subsection{Evaluation Metrics}
\label{sec:eval-metrics}

We utilize the following metrics to evaluate behavioral performance and internal epistemic trends. Detailed definitions and implementation specifications for each metric are provided in Appendix~\ref{sec:appendix-metrics}.

\textbf{Behavioral Metrics (Output $y$).}
These metrics evaluate observable performance and epistemic behavior from generated outputs, including 
\textit{Accuracy} (fraction of correct answers in factual tasks), 
\textit{Pass@$k$} (probability of obtaining at least one correct solution among $k$ samples for reasoning tasks), 
\textit{Decisional Consistency (DC)} (stability across $N$ stochastic generations), and 
\textit{Self-Verify} (the model’s ability to assess its own correctness via prompting).

\textbf{Probabilistic Confidence (PC).}
These metrics capture internal knowledge states through probabilistic preferences or hidden representations, including 
\textit{Normalized Probability} (length-normalized conditional likelihood of answers), 
\textit{Linear Probing} (linear separability of correctness from hidden states), 
\textit{MARS} (semantic uncertainty via meaning-weighted token probabilities; we report AUROC of the combined score), and 
\textit{SAPLMA} (activation-based prediction of latent correctness; reported for zero-shot only, as the method is prompt-invariant).

\textbf{Relative Ranking Analysis.}
Because these metrics capture different properties---such as behavioral correctness, uncertainty, and latent separability---their absolute values are not directly comparable. We therefore additionally report the relative \emph{rank} of each condition across the corresponding prompting settings (shown in parentheses), where lower ranks indicate stronger performance. These rankings allow us to analyze whether certain conditions are systematically up- or down-ranked by internal metrics relative to behavioral metrics such as Accuracy.

\section{Results}
\label{sec:05results}

\begin{table*}[t]
  \centering
  \caption{Head vs.\ Tail evaluation (with Gap) under standard prompting (zero-shot and few-shot).} 
  \label{tab:RQ2_head_tail_gap1}

  \begingroup
  \setlength{\tabcolsep}{3pt}
  \renewcommand{\arraystretch}{0.92}
  \scriptsize

    \setlength{\tabcolsep}{10pt}
  \resizebox{\textwidth}{!}{%
    \setlength{\tabcolsep}{15pt}
\renewcommand{\arraystretch}{0.95}
{\small
\begin{tabular}{
  >{\centering\arraybackslash}m{0.75cm}   
  >{\centering\arraybackslash}m{1.35cm}   
  l                       
  c c >{\columncolor{gray!8}}c          
  c c >{\columncolor{gray!8}}c          
}
\toprule
& & & \multicolumn{3}{c}{\textbf{Zero-shot}}
    & \multicolumn{3}{c}{\textbf{Few-shot (5)}} \\
\cmidrule(lr){4-6}\cmidrule(lr){7-9}

& & &
\textbf{Head} & \textbf{Tail} & \textbf{Gap}
& \textbf{Head} & \textbf{Tail} & \textbf{Gap} \\
\cmidrule(lr){4-4}\cmidrule(lr){5-5}\cmidrule(lr){6-6}
\cmidrule(lr){7-7}\cmidrule(lr){8-8}\cmidrule(lr){9-9}

\textbf{OLMo~2} & \textbf{Types} & \textbf{Metrics}
& & & 
& & & \\
\midrule

\multirow{7}{*}{\makecell{\textbf{7B}}}
& \multirow{3}{*}{\makecell{\textbf{Behavioral}}}
& Accuracy        & 86.54 (10) & \textbf{53.64 (23)} & 32.90 (-13) & 88.80 (5) & \textbf{63.28 (18)} & 25.52 (-13) \\
&  & $DC$     & \textbf{92.71 (11)} & \textbf{74.01 (23)} & 18.70 (-12) & 95.99 (5) & 81.92 (16) & 14.06 (-11) \\
&  & Self-Verify  & 25.42 (22) & 43.17 (16) & -17.75 (6) & 23.69 (24) & 35.84 (19) & -12.15 (5) \\
\cmidrule{2-3}
& \multirow{4}{*}{\makecell{\textbf{Internal}\\$(PC)$}}
& Norm. $P$ & 72.84 (8) & 42.14 (19) & 30.71 (-11) & 71.85 (9) & 41.79 (20) & 30.06 (-11) \\
&  & MARS            & 57.53 (23) & 62.46 (13) & -4.93 (10) & 61.39 (16) & 63.01 (12) & -1.62 (4) \\
&  & Linear Probing  & 82.33 (18) & \textbf{84.29 (9)} & -1.96 (9) & 83.20 (13) & 79.80 (21) & 3.40 (-8) \\
&  & SAPLMA          & 90.78 (3) & 75.70 (6) & 15.09 (-3) & - & - &  - \\
\addlinespace[2pt]
\midrule

\multirow{7}{*}{\makecell{\textbf{13B}}}
& \multirow{3}{*}{\makecell{\textbf{Behavioral}}}
& Accuracy        & 87.73 (8) & 60.92 (20) & 26.81 (-12) & 85.80 (11) & 64.32 (17) & 21.48 (-6) \\
&  & $DC$     & 94.38 (7) & 76.00 (22) & 18.38 (-15) & 94.11 (8) & 78.61 (20) & 15.50 (-12) \\
&  & Self-Verify  & 63.66 (11) & 47.73 (14) & 15.93 (-3) & 64.75 (10) & 49.61 (13) & 15.15 (-3) \\
\cmidrule{2-3}
& \multirow{4}{*}{\makecell{\textbf{Internal}\\$(PC)$}}
& Norm. $P$ & 32.46 (21) & 26.61 (23) & 5.85 (-2) & 32.32 (22) & 26.41 (24) & 5.91 (-2) \\
&  & MARS            & 62.41 (14) & 65.00 (8) & -2.59 (6) & 65.32 (6) & 66.62 (3) & -1.30 (3) \\
&  & Linear Probing  & 82.84 (15) & 82.21 (19) & 0.63 (-4) & 86.82 (6) & 80.27 (20) & 6.55 (-14) \\
&  & SAPLMA          & 91.34 (2) & 77.45 (5) & 13.89 (-3) & - & - & - \\
\addlinespace[2pt]
\midrule

\multirow{7}{*}{\makecell{\textbf{32B}}}
& \multirow{3}{*}{\makecell{\textbf{Behavioral}}}
& Accuracy        & 92.20 (1) & 68.23 (13) & 23.98 (-12) & 92.13 (2) & 67.71 (14) & 24.42 (-12) \\
&  & $DC$     & 96.98 (3) & 82.47 (15) & 14.51 (-12) & 96.44 (4) & 79.36 (18) & 17.09 (-14) \\
&  & Self-Verify  & 92.31 (2) & 67.04 (7) & 25.26 (-5) & 93.50 (1) & 67.56 (6) & 25.94 (-5) \\
\cmidrule{2-3}
& \multirow{4}{*}{\makecell{\textbf{Internal}\\$(PC)$}}
& Norm. $P$ & 60.19 (15) & 48.11 (18) & 12.08 (-3) & 60.08 (16) & 49.37 (17) & 10.71 (-1)  \\
&  & MARS            & 58.11 (21) & 66.78 (2) & -8.67 (19) & 57.94 (22) & 67.78 (1) & -9.84 (21) \\
&  & Linear Probing  & 72.29 (24) & 84.29 (9) & -12.00 (15) & 86.98 (5) & 83.63 (12) & 3.35 (-7) \\
&  & SAPLMA          & 93.02 (1) & 79.36 (4) & 13.66 (-3) & - & - & - \\
\bottomrule
\end{tabular}}

  }
  \endgroup
\end{table*}

Throughout our analysis, we prioritize \emph{same-metric, across-condition} comparisons (e.g., Seen vs.\ Unseen Accuracy), which are directly interpretable. When cross-metric comparison is necessary, we rely on relative rankings within each condition (shown in parentheses; lower is better), as our claims concern the \textit{direction and consistency} of the Seen/Unseen gap rather than absolute magnitudes. 


\subsection{Results on Factual Domain}

We evaluate under zero-shot and few-shot (5-shot) prompting;
prompt templates in Appendix~\ref{prompts:factual}.

\textbf{Models internally encode knowledge they fail to express, but this gap narrows with scale.}
Table~\ref{tab:RQ1_factual} confirms the expected exposure-driven pattern in Accuracy. However, internal metrics do not always follow the same ordering. At 7B zero-shot, Seen Web Linear Probing ranks 5th while Accuracy ranks 23rd, indicating that the latent space $\mathbf{h}$ encodes task-relevant information not reliably surfaced during decoding. This positive rank gap is consistently observed in Seen conditions at 7B and 13B but absent in Unseen conditions, revealing that the bottleneck arises from a failure to express already-encoded knowledge rather than from missing encoding---extending the expression gap of \citep{gekhman2025inside} with an exposure-aware diagnosis.
This resolution is progressive: at 13B, the Seen Web rank gap narrows to $+13$ (Linear Probing rank 6, Accuracy rank 19 in zero-shot), intermediate between the 7B gap of $+18$ and the reversed gap at 32B, where Accuracy rank 1 surpasses Linear Probing rank 23 in Seen Academic zero-shot.
The frequency dimension reinforces this finding. As shown in Table~\ref{tab:RQ2_head_tail_gap1}, Tail entities---appearing only once in the corpus---achieve Accuracy well above the 25\% baseline across all scales (53.64\% at 7B to 68.23\% at 32B), yet with markedly lower DC than Head counterparts (74.01\% vs.\ 92.71\% at 7B). Linear Probing confirms this is a retrieval failure, not an encoding absence: the Tail Linear Probing score (rank 9) far exceeds behavioral Accuracy (rank 23) at 7B. This ``recognition-over-recall'' pattern---where the model retains enough parametric residue to recognize facts in MCQ format but cannot stably produce them---directly supports the retrieval bottleneck hypothesis for low-frequency knowledge. This pattern persists under Chain-of-Thought prompting, where the Head/Tail gap remains structurally consistent (Appendix Table~\ref{sec:cot_head_tail_new}).

\begin{table*}[t]
  \centering
  \caption{Self-Verify analysis results under standard prompting (zero-shot and few-shot).}
  \label{tab:A_self_verify_cf}

  \begingroup
  \setlength{\tabcolsep}{3pt}
  \renewcommand{\arraystretch}{0.92}
  \scriptsize

    \setlength{\tabcolsep}{10pt}
  \resizebox{\textwidth}{!}{%
    \setlength{\tabcolsep}{6pt}
\renewcommand{\arraystretch}{0.8}
\resizebox{\textwidth}{!}{
\begin{tabular}{
  >{\centering\arraybackslash}m{1.1cm}
  l
  c c >{\columncolor{gray!8}}c >{\columncolor{gray!8}}c
  c c >{\columncolor{gray!8}}c >{\columncolor{gray!8}}c
}

\toprule
& & \multicolumn{4}{c}{\textbf{Zero-shot}}
  & \multicolumn{4}{c}{\textbf{Few-shot (5)}} \\
\cmidrule(lr){3-6} \cmidrule(lr){7-10}

& &
\multicolumn{2}{c}{Seen} &
\multicolumn{2}{c}{Unseen} &
\multicolumn{2}{c}{Seen} &
\multicolumn{2}{c}{Unseen} \\
\cmidrule(lr){3-4}
\cmidrule(lr){5-6}
\cmidrule(lr){7-8}
\cmidrule(lr){9-10}

\textbf{OLMo~2} & \textbf{Metrics}
& \makecell{\textbf{Academic}}
& \makecell{\textbf{Web}}
& \makecell{\textbf{RealTime}\\\textbf{QA}}
& \makecell{\textbf{Current}\\\textbf{events}}
& \makecell{\textbf{Academic}}
& \makecell{\textbf{Web}}
& \makecell{\textbf{RealTime}\\\textbf{QA}}
& \makecell{\textbf{Current}\\\textbf{events}} \\
\midrule

\multirow[c]{4}{*}{\makecell[c]{\textbf{7B}}}
& Accuracy        & \textbf{28.00} & 34.29 & 61.27 & \textbf{66.30} & 26.52 & 29.77 & 59.94 & 63.61 \\
& Precision            & 14.13 & 14.62 & 9.70 & \textbf{15.65} & 12.68 & 13.30 & 7.88 & 13.74 \\
& Recall     & 14.13 & 14.62 & 9.70 & \textbf{15.65} & 12.68 & 13.33 & 7.88 & 13.74 \\
& F1      & 14.13 & 14.62 & 9.70 & \textbf{15.65} & 12.68 & 13.31 & 7.88 & 13.74 \\

\midrule

\multirow[c]{4}{*}{\makecell[c]{\textbf{13B}}}
& Accuracy        & 62.54 & 55.70 & 49.53 & 48.42 & 60.89 & 57.18 & 49.53 & 48.02 \\
& Precision            & 56.64 & 50.32 & 32.08 & 35.21 & 55.90 & 51.31 & 30.61 & 33.20 \\
& Recall     & 61.45 & 57.38 & 43.64 & 48.85 & 61.81 & 56.81 & 41.21 & 45.80 \\
& F1       & 57.89 & 52.09 & 34.78 & 38.22 & 57.45 & 52.79 & 33.06 & 36.19 \\

\midrule

\multirow[c]{4}{*}{\makecell[c]{\textbf{32B}}}
& Accuracy        & \textbf{83.76} & 79.67 & \textbf{54.77} & \textbf{53.48} & 83.83 & 80.53 & 53.88 & 53.57 \\
& Precision        & 79.69 & 76.12 & 39.15 & 42.38 & 79.55 & 76.58 & 37.66 & 41.61 \\
& Recall     & 81.88 & 82.32 & 47.27 & 54.71 & 82.25 & 83.18 & 47.27 & 57.25 \\
& F1       & \textbf{80.22} & 77.80 & \underline{\textbf{41.11}} & \underline{\textbf{45.07}} & 80.34 & 78.45 & 39.98 & 45.21 \\

\bottomrule

\end{tabular}
}

  }
  \endgroup
\end{table*}

\begin{table*}[t]
  \centering
  \caption{Factual evaluation across Seen/Unseen splits under Chain-of-Thought prompting (zero-shot and few-shot).}
  \label{tab:RQ1_factual_cot}

  \begingroup
  \setlength{\tabcolsep}{3pt}
  \renewcommand{\arraystretch}{1.15}
  \small

  \resizebox{\textwidth}{!}{%
    \setlength{\tabcolsep}{6pt}
\renewcommand{\arraystretch}{1.15}
\resizebox{\textwidth}{!}{
\begin{tabular}{
  >{\centering\arraybackslash}m{1.6cm}
  >{\centering\arraybackslash}m{2.2cm}
  l
  c c >{\columncolor{gray!8}}c >{\columncolor{gray!8}}c
  c c >{\columncolor{gray!8}}c >{\columncolor{gray!8}}c
}

\toprule
& & & \multicolumn{4}{c}{\textbf{CoT Zero-shot}}
    & \multicolumn{4}{c}{\textbf{CoT Few-shot (5)}} \\
\cmidrule(lr){4-7} \cmidrule(lr){8-11}

& & &
\multicolumn{2}{c}{Seen} &
\multicolumn{2}{c}{Unseen} &
\multicolumn{2}{c}{Seen} &
\multicolumn{2}{c}{Unseen} \\
\cmidrule(lr){4-5}
\cmidrule(lr){6-7}
\cmidrule(lr){8-9}
\cmidrule(lr){10-11}

\textbf{\textbf{OLMo~2}} & \textbf{Types} & \textbf{Metrics}
& \makecell{\textbf{Academic}}
& \makecell{\textbf{Web}}
& \makecell{\textbf{RealTime}\\\textbf{QA}}
& \makecell{\textbf{Current}\\\textbf{events}}
& \makecell{\textbf{Academic}}
& \makecell{\textbf{Web}}
& \makecell{\textbf{RealTime}\\\textbf{QA}}
& \makecell{\textbf{Current}\\\textbf{events}} \\
\midrule

\multirow{6}{*}{\makecell{\textbf{7B}}}
& \multirow{3}{*}{\makecell{\textbf{Behavioral}}}
& Accuracy        & 71.25 (22) & 66.34 (24) & \underline{31.74 (48)} & 32.01 (47) & 75.63 (15) & 74.10 (21) & 35.10 (44) & 35.58 (42) \\
&  & $DC$     & 83.09 (24) & 80.22 (28) & 66.62 (48) & 68.46 (47) & 89.47 (12) & 85.92 (20) & 73.10 (40) & 74.60 (35) \\
&  & Self-Verify  & 31.95 (43) & 36.79 (41) & 62.09 (13) & 66.41 (9) & 26.09 (48) & 29.94 (44) & 55.21 (22) & 61.35 (14) \\
\cmidrule{2-3}
& \multirow{4}{*}{\makecell{\textbf{Internal}\\($PC$)}}
& Norm. $P$ & 73.69 (4) & 73.53 (5) & \underline{52.83 (27)} & 47.88 (28) & 74.82 (3) & 69.37 (11) & 54.06 (25) & 53.45 (26) \\
&  & MARS            & 59.84 (19) & 63.29 (5) & 56.37 (36) & 62.59 (8) & 59.00 (23) & 63.26 (6) & 57.23 (32) & 60.61 (15) \\
&  & Linear Probing  & 80.53 (9) & 86.45 (1) & 55.77 (36) & 69.88 (26) & 75.37 (20) & 79.58 (13) & 54.01 (42) & 70.42 (24) \\
\addlinespace[2pt]
\midrule

\multirow{6}{*}{\makecell{\textbf{13B}}}
& \multirow{3}{*}{\makecell{\textbf{Behavioral}}}
& Accuracy        & 79.78 (7) & 75.63 (15) & 38.18 (36) & 38.40 (35) & 78.69 (13) & 74.11 (20) & 42.53 (33) & 42.01 (34) \\
&  & $DC$     & 89.41 (13) & 86.48 (18) & 68.84 (46) & 73.80 (37) & 91.32 (6) & 87.72 (16) & 73.50 (39) & 78.29 (33) \\
&  & Self-Verify  & 38.20 (40) & 40.73 (39) & 52.29 (28) & 53.23 (27) & 60.14 (18) & 61.13 (16) & 49.04 (34) & 47.19 (38) \\
\cmidrule{2-3}
& \multirow{4}{*}{\makecell{\textbf{Internal}\\($PC$)}}
& Norm. $P$ & 70.68 (8) & 73.40 (6) & 60.74 (17) & 55.46 (22) & 81.01 (2) & 83.46 (1) & 71.00 (7) & 63.02 (16) \\
&  & MARS            & 59.41 (22) & 62.52 (9) & 57.90 (29) & 58.48 (25) & 57.24 (31) & 62.46 (11) & 56.74 (34) & 59.82 (20) \\
&  & Linear Probing  & 78.39 (14) & 85.04 (3) & 57.87 (35) & 59.06 (34) & 79.71 (12) & 71.71 (22) & 50.80 (45) & 63.99 (30) \\
\addlinespace[2pt]
\midrule

\multirow{6}{*}{\makecell{\textbf{32B}}}
& \multirow{3}{*}{\makecell{\textbf{Behavioral}}}
& Accuracy        & 82.41 (2) & 78.93 (12) & 45.84 (25) & 44.32 (27) & 82.18 (4) & 79.30 (9) & 45.12 (26) & 43.58 (30) \\
&  & $DC$     & 94.41 (2) & 90.90 (9) & 83.19 (23) & 82.29 (25) & 93.97 (3) & 90.95 (8) & 81.40 (26) & 81.27 (27) \\
&  & Self-Verify  & 83.56 (3) & 79.12 (8) & 48.98 (35) & 49.87 (31) & 82.28 (4) & 79.81 (6) & 50.91 (29) & 50.27 (30) \\
\cmidrule{2-3}
& \multirow{4}{*}{\makecell{\textbf{Internal}\\($PC$)}}
& Norm. $P$ & 68.14 (12) & 67.27 (13) & 41.85 (32) & 44.45 (30) & 69.80 (10) & 70.02 (9) & 42.22 (31) & 47.20 (29) \\
&  & MARS            & 56.19 (37) & 62.50 (10) & 52.83 (42) & 59.82 (20) & 54.84 (40) & 61.03 (14) & 55.78 (39) & 56.52 (35) \\
&  & Linear Probing  & 68.78 (27) & 77.80 (16) & 55.67 (37) & 68.66 (28) & 72.55 (21) & 79.88 (11) & 55.64 (38) & 54.39 (41) \\

\bottomrule
\end{tabular}}

  }
  
  \endgroup
\end{table*}

\textbf{Meta-cognitive calibration collapses at the exposure boundary, and CoT does not resolve it.}
Self-Verify Accuracy rises from 28.00\% (7B) to 83.76\% (32B) on Seen Academic, suggesting a positive scaling trend consistent with \citep{mahaut2024factual}'s finding that verbalization-based confidence is only competitive for sufficiently capable models. However, a confusion matrix analysis (Table~\ref{tab:A_self_verify_cf}) reveals a more complex picture. At 7B, the model's seemingly higher Unseen accuracy (e.g., 66.30\% on Current Events) is undermined by precision and recall both below 16\%, confirming indiscriminate overconfidence. At 32B, Seen self-verify reaches 83.76\% accuracy with F1 of 80.22\%, demonstrating genuine meta-cognitive ability---yet this ability collapses in Unseen domains, where accuracy hovers near the 50\% baseline (binary-classification) and F1 drops to 41--45\%. 
This collapse specifically emerges at the seen/unseen boundary rather than being a fixed property of model scale.

Chain-of-Thought prompting does not resolve this failure. As shown in Table~\ref{tab:A_self_verify_cf_cot},
\textcolor{lumosblue}{CoT modestly improves 32B Unseen F1 (41.11\% to 46.73\% on RealTimeQA; 45.07\% to 48.12\% on Current Events), but the scores remain substantially below Seen Academic (83.37\%).}
More critically, CoT inflates Normalized $P$ without corresponding accuracy gains: at 7B, Unseen RealTimeQA Norm.\ $P$ rises sharply from 28.18\% to 52.83\% under CoT (Table~\ref{tab:RQ1_factual_cot}), yet Accuracy slightly \textit{decreases} (33.88\% to 31.74\%). This dissociation between elevated confidence and stagnant performance indicates that CoT produces diagnostically misleading signals rather than genuine calibration improvement.

\textcolor{lumosblue}{\textbf{Exposure-conditioned performance gaps persist despite scaling.}}
Across all scales, a persistent 30--40 percentage point gap separates Seen and Unseen Accuracy, with Unseen performance remaining below 44\% even at 32B.
This gap persists regardless of prompting strategy: few-shot examples improve Seen Web Accuracy from 70.09\% to 76.04\% at 7B but yield negligible gains on Unseen RealTimeQA (33.88\% to 35.70\%), and this marginal benefit diminishes further at 32B. This confirms that in-context learning acts as a retrieval trigger for already-encoded knowledge---narrowing the $h \to y$ gap---but cannot compensate for the absence of exposure in $\mathcal{D}$.
\textcolor{lumosblue}{Similar exposure-conditioned differences}
extend beyond Accuracy to internal metrics: at 32B, $DC$ maintains a persistent Seen/Unseen gap (94.73\% vs.\ 78.73\% in zero-shot), and Linear Probing similarly shows higher separability for Seen conditions (78.29\% vs.\ 50.55\% for Web/RealTimeQA), confirming that the exposure boundary constrains not only behavioral output but also the model's internal epistemic organization.
\textcolor{lumosblue}{Importantly, a Seen/Unseen accuracy gap also persists on the cutoff-only unseen set (Appendix~\ref{sec:cutoff_only}), although its magnitude varies. Together, these observations highlight the importance of training exposure in factual reliability across model scales.}

\begin{table*}[t]
  \centering
  \caption{Self-Verify analysis results under Chain-of-Thought prompting.}
  \label{tab:A_self_verify_cf_cot}

  \begingroup
  \setlength{\tabcolsep}{3pt}
  \renewcommand{\arraystretch}{0.92}
  \scriptsize

    \setlength{\tabcolsep}{10pt}
  \resizebox{\textwidth}{!}{%
    \setlength{\tabcolsep}{6pt}
\renewcommand{\arraystretch}{0.8}
\resizebox{\textwidth}{!}{
\begin{tabular}{
  >{\centering\arraybackslash}m{1.1cm}
  l
  c c >{\columncolor{gray!8}}c >{\columncolor{gray!8}}c
  c c >{\columncolor{gray!8}}c >{\columncolor{gray!8}}c
}

\toprule
& & \multicolumn{4}{c}{\textbf{CoT Zero-shot}}
  & \multicolumn{4}{c}{\textbf{CoT Few-shot (5)}} \\
\cmidrule(lr){3-6} \cmidrule(lr){7-10}

& &
\multicolumn{2}{c}{Seen} &
\multicolumn{2}{c}{Unseen} &
\multicolumn{2}{c}{Seen} &
\multicolumn{2}{c}{Unseen} \\
\cmidrule(lr){3-4}
\cmidrule(lr){5-6}
\cmidrule(lr){7-8}
\cmidrule(lr){9-10}

\textbf{OLMo~2} & \textbf{Metrics}
& \makecell{\textbf{Academic}}
& \makecell{\textbf{Web}}
& \makecell{\textbf{RealTime}\\\textbf{QA}}
& \makecell{\textbf{Current}\\\textbf{events}}
& \makecell{\textbf{Academic}}
& \makecell{\textbf{Web}}
& \makecell{\textbf{RealTime}\\\textbf{QA}}
& \makecell{\textbf{Current}\\\textbf{events}} \\
\midrule

\multirow{4}{*}{\makecell{\textbf{7B}}}
& Accuracy        & 31.95 & 36.79 & 62.09 & 66.41 & 26.09 & 29.94 & 55.21 & 61.35 \\
& Precision            & 12.70 & 15.69 & 13.21 & 18.24 & 18.57 & 22.02 & 18.90 & 22.03 \\
& Recall     & 12.46 & 14.69 & 12.81 & 18.10 & 13.33 & 16.22 & 14.84 & 20.86 \\
& F1      & 12.47 & 14.70 & 12.81 & 17.68 & 14.13 & 16.95 & 15.89 & 21.04 \\

\midrule

\multirow{4}{*}{\makecell{\textbf{13B}}}
& Accuracy        & 38.20 & 40.73 & 52.29 & 53.23 & 60.14 & 61.13 & 49.04 & 47.19 \\
& Precision            & 52.33 & 53.01 & 36.10 & 40.52 & 73.23 & 70.82 & 44.78 & 42.74 \\
& Recall     & 32.68 & 38.54 & 41.30 & 48.30 & 62.63 & 69.63 & 65.15 & 61.05 \\
& F1       & 35.66 & 39.39 & 33.99 & 39.77 & 63.01 & 65.11 & 49.08 & 46.39 \\

\midrule

\multirow{4}{*}{\makecell{\textbf{32B}}}
& Accuracy        & 83.56 & 79.12 & 48.98 & 49.87 & 82.28 & 79.81 & 50.91 & 50.27 \\
& Precision        & 82.62 & 79.01 & 45.71& 45.67 & 82.73 & 79.63 & 43.64 & 45.05 \\
& Recall     & 86.93 & 86.82 & 56.81 & 61.81 & 85.14 & 85.63 & 50.66 & 60.48 \\
& F1       & \textbf{83.37} & 80.29 & \underline{\textbf{46.73}} & \underline{\textbf{48.12}} & 82.44 & 80.17 & 44.11 & 47.38 \\

\bottomrule

\end{tabular}

}
  }

  \endgroup
\end{table*}

\subsection{Generalization in Reasoning}

\textbf{RLVR and SFT generalize through qualitatively different mechanisms, and their relative advantage shifts with scale.}

Table~\ref{tab:RQ1_reasoning} summarizes reasoning performance across scales and training methods. Unlike the factual domain, reasoning performance is governed by an interplay between task difficulty and learned representations.
At 7B, SFT exhibits brittle pattern matching: Seen MetaMath (20.06\% Pass@1) outperforms all Unseen datasets (GSM8K-Test 17.75\%, GSM1K 18.27\%, GSM-Symbolic 15.73\%), suggesting a memorization buffer. This advantage collapses at 13B, where Unseen GSM8K-Test (64.90\%) first surpasses Seen MetaMath (59.80\%), and widens further at 32B (75.55\% vs.\ 63.92\%)---indicating that SFT overcomes its memorization bias only as model capacity grows.
RLVR shows this pattern from the start: even at 7B, Unseen GSM8K-Test (74.27\%) already outperforms Seen MetaMath (66.56\%), because reinforcement learning optimizes for reward-aligned reasoning procedures rather than surface-level templates. GSM-Symbolic corroborates this: at 32B, RLVR maintains a narrow gap between GSM8K-Test and GSM-Symbolic (87.64\% vs.\ 83.37\%), whereas SFT shows greater sensitivity to numerical perturbation (75.55\% vs.\ 67.14\%). However, the RLVR advantage narrows substantially at 32B, suggesting that at smaller scales RLVR builds genuine reasoning capability, while at larger scales its advantage reduces to sampling efficiency. Moreover, this pattern is domain-dependent: as shown in Appendix~\ref{sec:Results_in_Code_Dataset}, RLVR's gains diminish in code generation, where reward-task alignment is weaker.
Taken together, both methods converge on the same conclusion---task difficulty supersedes training exposure in reasoning---but via different mechanisms: SFT reaches it only as capacity overcomes memorization bias, while RLVR embeds it from the start.

\begin{table}[t!]
  \centering
  \caption{Reasoning evaluation across Seen/Unseen splits and model variants.}
  \label{tab:RQ1_reasoning}
  \vspace{8pt}

  \setlength{\tabcolsep}{5pt}
  \renewcommand{\arraystretch}{1.18}
  \footnotesize

  \resizebox{\columnwidth}{!}{%
\normalsize
\setlength{\tabcolsep}{3pt}
\renewcommand{\arraystretch}{1.25}
\resizebox{\columnwidth}{!}{
\begin{tabular}{
  l l
  @{\hspace{1.5em}} cc >{\columncolor{gray!8}}c >{\columncolor{gray!8}}c >{\columncolor{gray!8}}c
  @{\hspace{1.5em}} cc >{\columncolor{gray!8}}c >{\columncolor{gray!8}}c >{\columncolor{gray!8}}c
  @{\hspace{1.5em}} cc >{\columncolor{gray!8}}c >{\columncolor{gray!8}}c >{\columncolor{gray!8}}c
}
\toprule
\multicolumn{2}{c}{} 
& \multicolumn{5}{c}{\textbf{7B}}
& \multicolumn{5}{c}{\textbf{13B}}
& \multicolumn{5}{c}{\textbf{32B}} \\
\cmidrule(lr){3-7} \cmidrule(lr){8-12} \cmidrule(lr){13-17}

\multicolumn{2}{c}{} 
& \multicolumn{2}{c}{\textbf{Seen}}
& \multicolumn{3}{c}{\textbf{Unseen}}
& \multicolumn{2}{c}{\textbf{Seen}}
& \multicolumn{3}{c}{\textbf{Unseen}}
& \multicolumn{2}{c}{\textbf{Seen}}
& \multicolumn{3}{c}{\textbf{Unseen}} \\

\cmidrule(lr){3-4}  \cmidrule(lr){5-7}
\cmidrule(lr){8-9}  \cmidrule(lr){10-12}
\cmidrule(lr){13-14}\cmidrule(lr){15-17}

\textbf{Method} & 
& \makecell{\textbf{G8K}\\\textbf{Train}}
& \textbf{MMath}
& \makecell{\textbf{G8K}\\\textbf{Test}}
& \textbf{G1K}
& \textbf{G-Sym}
& \makecell{\textbf{G8K}\\\textbf{Train}}
& \textbf{MMath}
& \makecell{\textbf{G8K}\\\textbf{Test}}
& \textbf{G1K}
& \textbf{G-Sym}
& \makecell{\textbf{G8K}\\\textbf{Train}}
& \textbf{MMath}
& \makecell{\textbf{G8K}\\\textbf{Test}}
& \textbf{G1K}
& \textbf{G-Sym} \\

\midrule

\multirow{4}{*}{\textbf{SFT}} 
& $P$@1 & 21.48 & \textbf{20.06} & \textbf{17.75} & \textbf{18.27} & \textbf{15.73}
        & 70.73 & \textbf{59.80} & \textbf{64.90} & 57.95 & 59.51
        & 79.88 & \textbf{63.92} & \textbf{75.55} & 69.65 & \textbf{67.14} \\
& $P$@2 & 37.10 & 34.14 & 31.22 & 31.48 & 27.84
        & 86.43 & 74.84 & 81.30 & 74.50 & 75.97
        & 91.86 & 78.18 & 88.51 & 83.22 & 81.36 \\
& $P$@4 & 57.41 & 52.06 & 49.79 & 48.85 & 44.76
        & 93.64 & 83.53 & 90.56 & 84.98 & 86.26
        & 96.26 & 85.85 & 94.02 & 89.85 & 89.38 \\
& $P$@8 & 77.33 & 69.40 & 69.07 & 66.14 & 62.80
        & 96.59 & 88.40 & 95.15 & 90.62 & 91.50
        & 97.65 & 89.70 & 96.29 & 93.44 & 93.40 \\
\midrule

\multirow{4}{*}{\textbf{RLVR}} 
& $P$@1 & 80.18 & \textbf{66.56} & \textbf{74.27} & 66.56 & 65.74
        & 85.73 & 72.86 & 81.02 & 74.85 & 76.46
        & 90.12 & 71.98 & \textbf{87.64} & 82.38 & \textbf{83.37} \\
& $P$@2 & 90.52 & 78.77 & 85.66 & 78.77 & 77.96
        & 93.03 & 81.29 & 89.49 & 85.03 & 86.02
        & 95.42 & 82.06 & 93.33 & 89.63 & 90.53 \\
& $P$@4 & 94.70 & 85.38 & 91.33 & 85.78 & 85.13
        & 95.96 & 85.74 & 93.41 & 90.47 & 90.83
        & 96.94 & 87.35 & 95.28 & 93.04 & 93.55 \\
& $P$@8 & 96.59 & 89.20 & 94.31 & 89.88 & 89.30
        & 97.35 & 88.50 & 95.38 & 93.61 & 93.40
        & 97.65 & 90.20 & 96.51 & 95.10 & 95.40 \\
\bottomrule
\end{tabular}
}
  }
\end{table}
\section{Conclusion}
\label{sec:06conclusion}

We introduced \method, a diagnostic framework that grounds LLM knowledge evaluation in verified training-data exposure. By tracing knowledge from pretraining corpus to behavioral output, we show that training exposure
\textcolor{lumosblue}{plays a central role in}
factual reliability---a structural constraint that scaling, prompting, and Chain-of-Thought reasoning cannot overcome. We advocate for the training-data axis as a standard component of LLM knowledge assessment.




\section{Limitations}
\label{sec:limitations}

Our study has several limitations. First, our analysis relies on models with fully accessible pretraining corpora, limiting the direct applicability of \method~to closed-source or proprietary LLMs. Moreover, our definition of training exposure relies on explicit string matching and frequency statistics, which may overlook implicit, paraphrased, or indirectly encoded knowledge.
\textcolor{lumosblue}{As our experiments focus on a single model family, extending the analysis to other fully open-source LLMs would help validate the generalizability of our findings. Additionally, while we report precision, recall, and F1 scores for self-verification, we do not explicitly account for class balance or examine how the proportions of CORRECT and INCORRECT judgments vary across model scales and exposure conditions. Further analysis of these distributions could provide a more nuanced interpretation of self-verification performance. Beyond the prompting-based comparisons of final-answer behavior conducted here, LUMOS could also facilitate future investigations into whether generated CoT reasoning causally contributes to final predictions.}
\section*{Acknowledgements}
This research was supported by grants from the National Research Foundation of Korea (NRF), funded by the Korean government (MSIT and MOE: RS-2025-16063688; MSIT: RS-2025-02215813, RS-2026-25491306, RS-2025-25442252); and by the Ewha Global Excellence Program Research Grant. \textcolor{lumosblue}{The authors declare no competing interests.}

\bibliographystyle{unsrtnat}
\bibliography{refs}

\appendix

\section{Experimental details}
\label{sec:experimental-details}

All experiments were conducted on a machine equipped with 4× NVIDIA RTX PRO 6000 GPUs (approximately 98GB total VRAM) running CUDA 13.0. For each question, we generate $K = 11$ responses per query---one deterministic output via greedy decoding and ten stochastic samples at temperature 1.0---which are used uniformly across all behavioral and internal metrics.

\begin{figure*}[h]
    \centering
    \includegraphics[width=\textwidth]{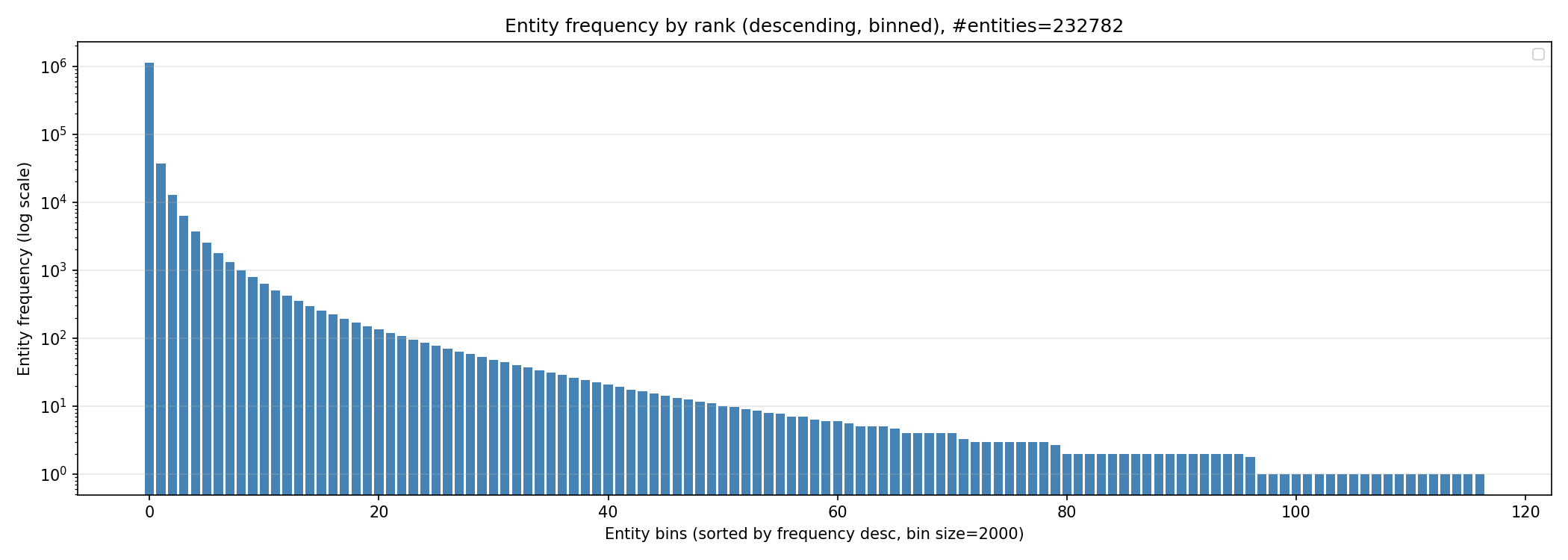}
    \caption{Entity frequency distribution under head--tail stratification.
    }
    \label{fig:entity-frequency}
\end{figure*}

\section{Evaluation Metrics}
\label{sec:appendix-metrics}

\subsection{Accuracy}
\label{sec:appendix-accuracy}

Accuracy measures the expected correctness of model outputs under stochastic decoding for a given multiple-choice question. Each generated answer is assigned a binary correctness indicator of $1$ if it matches the ground-truth answer and $0$ otherwise.

Since model generation is stochastic, we estimate accuracy by aggregating correctness over multiple generations per question. Specifically, for each question $q_i$, we obtain one deterministic (greedy) output and ten additional responses sampled with temperature $\text{temp}=1$, yielding a total of $K=11$ generated answers. We treat all generations uniformly when estimating accuracy.

Let $\hat{a}_i^{(k)}$ denote the $k$-th generated answer for question $q_i$, and let $a_i^*$ denote the ground-truth answer option. The per-question accuracy is defined as the fraction of correct responses among the $K$ generations:
\begin{equation}
\label{eq:per_question_accuracy}
\text{Acc}(q_i) = \frac{1}{K} \sum_{k=1}^{K} \mathbb{I}\bigl[\hat{a}_i^{(k)} = a_i^*\bigr].
\end{equation}

This formulation estimates the expected accuracy under the model’s output distribution and differs from majority-vote accuracy, as it evaluates all sampled responses rather than aggregating them into a single prediction.

Given the evaluation dataset $\mathcal{D} = \{(q_i, a_i^*)\}_{i=1}^{N}$, the dataset-level accuracy is computed by averaging the per-question accuracy across all questions:
\begin{equation}
\label{eq:dataset_accuracy}
\text{Accuracy} = \frac{1}{N} \sum_{i=1}^{N} \text{Acc}(q_i).
\end{equation}

\subsection{Decisional Consistency ($DC$)}

To measure how consistently a model selects the same answer across multiple stochastic generations, we define \emph{Decisional Consistency ($DC$)} as the degree of agreement among predictions for a given question.

Specifically, for each question \(q_i\), we obtain one deterministic (greedy) output and ten additional responses sampled with temperature \(\text{temp}=1\), resulting in \(K=11\) total predictions over the choice set \(\mathcal{A}=\{A,B,C,D\}\). We treat all generations uniformly when computing consistency.

Let \(n_i(a)\) denote the number of times option \(a \in \mathcal{A}\) is predicted for question \(q_i\). The per-question consistency is defined as the normalized frequency of the most frequently predicted answer:
\begin{equation}
\label{eq:per_question_consistency}
\text{Cons}(q_i) = \frac{1}{K}\max_{a \in \mathcal{A}} n_i(a).
\end{equation}

This metric estimates the concentration of the model’s predictive distribution over answer choices and is independent of correctness: it measures agreement among predictions regardless of whether the chosen answer is correct.

Given the evaluation dataset \(\mathcal{D} = \{(q_i, a_i^*)\}_{i=1}^{N}\), the dataset-level consistency is computed by averaging over questions:
\begin{equation}
\label{eq:dataset_consistency}
\text{Consistency} = \frac{1}{N}\sum_{i=1}^{N} \text{Cons}(q_i).
\end{equation}

\subsection{Normalized Probability}

Output probability score measures a length-normalized likelihood of a candidate answer given a question \(q\), thereby capturing the model’s relative preference for that answer. While accuracy evaluates only the binary correctness of the final output, this metric reflects how strongly the model favors a particular answer under its token-level distribution.

The conditional likelihood of an answer \(a = (a_1, \dots, a_n)\) is defined as follows. To reduce bias toward shorter or longer answers, we employ a length-normalized probability:

\begin{equation}
\label{eq:normalized_prob}
\begin{aligned}
P_{\text{norm}}(a \mid q)
&= \left( \prod_{i=1}^n P(a_i \mid q, a_{<i}) \right)^{\frac{1}{n}} \\
&= \exp\left( \frac{1}{n} \sum_{i=1}^n \log P(a_i \mid q, a_{<i}) \right)
\end{aligned}
\end{equation}

All probabilities are computed using teacher forcing, where the question and answer are concatenated into a single sequence. For reasoning-based QA, the probability is computed over the concatenated sequence \((q, r, a)\), where \(r\) denotes the model-generated reasoning.

Given a dataset \(\mathcal{D} = \{(q_i, a_i^*)\}_{i=1}^N\), we compute \(P_\text{norm}(a_i^* \mid q_i)\) for each sample and report the average normalized probability across the dataset.






\subsection{Linear Probing}
We evaluate the extent to which a language model's hidden states 
linearly encode information about answer correctness. This metric 
measures knowledge retention at the level of internal representations, 
independent of the model's final output.

For each question, we construct exactly one positive and one negative 
example as follows.
\begin{itemize}
    \item Positive: $(q, a^+, y=1)$, where $a^+$ is a correct answer 
    to $q$.
    \item Negative: $(q, a^-, y=0)$, where $a^-$ is an incorrect answer 
    sampled from the model's generations. If no incorrect generation is 
    available, $a^-$ is sampled uniformly from the remaining answer 
    choices.
\end{itemize}

Given the input sequence $\text{Prompt}(q) + a$, we extract the hidden 
state of the last non-padding token $h^{(l)}(q,a) \in \mathbb{R}^d$ 
from each layer $l \in \{1, \dots, L\}$ of the model. For each layer, 
we train an independent linear classifier (logistic regression with 
standardized features) to predict answer correctness from the 
corresponding hidden representation.

We select the best-performing layer based on AUROC on the development 
set. Using the probe 
trained on this selected layer, we compute a probe score 
$\hat{p} = P(y=1 \mid h^{(l^*)}(q, a)) \in [0,1]$ for each 
question--answer pair, where $l^*$ denotes the selected layer. We 
report the average probe score over the correct answers in the 
test set.

\subsection{Meaning-Aware Response Scoring (MARS)}
To more accurately quantify model uncertainty in generative settings beyond simple length-normalized likelihoods, we adopt the Meaning-Aware Response Scoring (MARS) framework proposed by~\cite{bakman2024marsmeaningawareresponsescoring}. Unlike conventional approaches that assign equal importance to all tokens, MARS incorporates the semantic contribution of each token to the overall meaning of the response in the context of the input question.

For a generated answer \( s = (s_1, \dots, s_L) \) to a question \( x \), the MARS score is defined as
\begin{equation}
\label{eq:mars_score}
\bar{P}(\mathbf{s} \mid \mathbf{x}, \theta)
= \prod_{l=1}^{L} P(s_l \mid s_{<l}, \mathbf{x}; \theta)^{w(\mathbf{s}, \mathbf{x}, L, l)},
\end{equation}
where \( P(s_l \mid s_{<l}, x; \theta) \) denotes the model’s conditional token probability, and \( w(\mathbf{s}, \mathbf{x}, L, l) \) is a weighting function that reflects both sequence length and semantic importance.

Specifically, the weight assigned to each token is defined as a convex combination of a uniform length-normalization term and a learned importance coefficient:
\begin{equation}
w(\mathbf{s}, \mathbf{x}, L, l)
\triangleq \frac{1}{2L} + \frac{u(\mathbf{s}, \mathbf{x}, l)}{2},
\end{equation}
where \( u(\mathbf{s}, \mathbf{x}, l) \in [0,1] \) represents the semantic importance of token \( s_l \) in the context of the question and the full generated sequence, and satisfies \( \sum_{l=1}^{L} u(\mathbf{s}, \mathbf{x}, l) = 1 \).

The importance coefficients \( u(\cdot) \) are computed by measuring the semantic impact of removing parts of the generated response. Concretely, MARS employs a BERT-based answer evaluation model that takes as input the question, the ground-truth answer, and the generated response, and predicts answer correctness. The importance of each token is estimated by evaluating the change in this correctness score when the token (or, in practice, a phrase containing the token) is masked. To account for dependencies between tokens, masking is performed at the phrase level rather than independently for each token. The resulting raw importance scores are then normalized (e.g., via softmax with temperature) to obtain \( u(\cdot) \).

MARS serves as a drop-in replacement for length-normalized scoring in probability-based uncertainty estimation methods, enabling these methods to better reflect the semantic relevance of generated tokens while preserving robustness to sequence length.

\subsection{Statement Accuracy Prediction based on Language Model Activations (SAPLMA)}
We adopt Statement Accuracy Prediction based on Language Model Activations (SAPLMA)~\cite{azaria2023internal}, a representation-based approach that predicts the truthfulness of a statement from a model’s internal hidden states.

Given an input statement, SAPLMA extracts hidden activations from a pretrained language model and feeds them into a lightweight classifier that outputs a scalar truthfulness score. Unlike probability-based metrics, this approach leverages internal representations that encode signals about factual correctness.

In our implementation, we perform a layer sweep and select the layer achieving the highest AUROC on a held-out development set, and train the classifier on the corresponding hidden representations.

We further adapt SAPLMA to the MCQ setting by evaluating all answer choices (A--D). For each question, we construct inputs of the form \texttt{[question + choice\_text]} without additional prompting, in order to isolate representation-level signals of parametric knowledge. We extract the hidden state at the final token for each input, and the classifier assigns a $p_{\text{true}}$ score to each choice. The highest-scoring choice is selected as the prediction, and we report performance as rank accuracy.

\subsection{Self-verify}
Self-verify measures how accurately a language model assesses the correctness of a proposed answer, capturing its meta-cognitive ability to distinguish correct from incorrect responses. Given a question $q$, reasoning $r$ and a candidate answer $a$, we prompt the model to explicitly judge whether the answer is correct, producing a binary decision (\emph{correct} or \emph{incorrect}). Under chain-of-thought prompting, the verification is conditioned on the triplet $(q, a, r)$, where $r$ denotes the generated reasoning.

\begin{center}
\begin{tcolorbox}[
  colback=gray!5,
  colframe=gray!60,
  title=Self-verify Evaluation Prompt,
  fonttitle=\bfseries,
  enhanced,
  breakable,
  boxrule=0.8pt,
  arc=3pt
]
\noindent
\textbf{System:} \\
Your job is to judge whether a proposed answer to a multiple-choice question is correct. \\

\vspace{1em}
\noindent
\textbf{User:} \\
Question: \texttt{\{question\}} \\
Choices: \texttt{\{choice\_text\}} \\
Reasoning: \texttt{\{reasoning\}} \\
Proposed Answer: \texttt{\{answer\_text\}} \\
\\
Is the proposed answer: \\
A: CORRECT \\
B: INCORRECT \\

\vspace{0.6em}
\noindent
Just return the letter A or B, with no text around it. \\
\end{tcolorbox}
\end{center}

Evaluation is performed by comparing the model’s judgment with the ground-truth correctness of the answer, yielding four possible outcomes: (i) answers that are correct and judged as correct, (ii) answers that are incorrect but judged as correct, (iii) answers that are correct but judged as incorrect, and (iv) answers that are incorrect and judged as incorrect. These outcomes together form a confusion matrix over all evaluated question--answer pairs.

Based on this confusion matrix, we compute standard classification metrics to quantify self-verify, including accuracy, precision, recall, and F1 score. In particular, self-verify accuracy is defined as the proportion of cases in which the model’s judgment matches the ground truth, i.e., the sum of cases (i) and (iv), normalized by the total number of evaluations.

For consistency, self-verify is evaluated on the same data splits used for accuracy and output-based metrics, enabling direct comparison across different measures of model knowledge.

\subsection{Pass@$k$}
Pass@$k$ is a standard metric for evaluating stochastic generation models in mathematical reasoning, measuring whether a correct answer appears among multiple sampled responses. Unlike single-sample accuracy, Pass@$k$ captures the model’s ability to recover a correct solution given multiple independent generation attempts.

Formally, for a given question, suppose the model generates $n$ responses in total, among which $c$ are correct. Pass@$k$ is defined as the expected probability that at least one correct response is observed when sampling $k$ responses without replacement:
\begin{equation}
\label{eq:passk}
\text{Pass@}k
=
\mathbb{E}\left[1 - \frac{\binom{n-c}{k}}{\binom{n}{k}}\right],
\end{equation}
where the expectation is taken over the stochastic generation process.

At the dataset level, Pass@$k$ is computed by averaging this quantity across all questions. While Pass@$k$ effectively measures coverage of correctness under sampling, it does not account for the diversity of correct solutions or the distribution of probability mass across responses.


\section{Prompts}
\label{sec:prompts}

\subsection{Unseen Filtering Prompt}

\begin{tcolorbox}[
  colback=gray!5,
  colframe=gray!60,
  title=Open-ended Short-answer Template,
  fonttitle=\bfseries,
  enhanced,
  breakable,
  boxrule=0.8pt,
  arc=3pt,
  center,
  before skip=4pt,
  after skip=4pt
]
\label{prompts:filtering}
\noindent
\textbf{System:}

\vspace{0.4em}
You are a concise assistant. Answer the question directly.

\vspace{1em}
\noindent
\textbf{User:}

\vspace{0.4em}
Question: \texttt{\{question\}}

\vspace{0.8em}
Answer concisely using a single word, a number, or a short phrase. \\
Do not answer in a complete sentence.
\end{tcolorbox}

\subsection{Factual Domain Prompt}

\label{prompts:factual}
\begin{tcolorbox}[
  colback=gray!5,
  colframe=gray!60,
  title=Zero-shot Prompt,
  fonttitle=\bfseries,
  enhanced,
  breakable,
  boxrule=0.8pt,
  arc=3pt,
  center,
  before skip=4pt,
  after skip=4pt
]
\noindent
\textbf{System:} 

\vspace{0.4em}
You are given a multiple-choice question. \\
You must answer \emph{only one letter}: A, B, C, or D. \\
Do not output anything else.

\vspace{1em}
\noindent
\textbf{User:} 

\vspace{0.4em}
Question: \texttt{\{question\}} \\
Choices: \texttt{\{choice\_text\}}

\vspace{0.8em}
\noindent
Answer with only one letter (A, B, C, or D).
\end{tcolorbox}


\begin{tcolorbox}[
  colback=gray!5,
  colframe=gray!60,
  title=Few-shot Prompt (5-shot),
  fonttitle=\bfseries,
  enhanced,
  breakable,
  boxrule=0.8pt,
  arc=3pt
]

\noindent
\textbf{System:} 

\vspace{0.4em}
You are given a multiple-choice question. \\
You must answer \emph{only one letter}: A, B, C, or D. \\
Do not output anything else.

\vspace{1em}
\noindent
\textbf{User:}

\vspace{0.4em}
\noindent
Question: What is the relationship between Carmen Miranda and Amaro Miranda da Cunha? \\
Choices: A. Siblings, B. Aunt and nephew, C. Mother and son, D. Cousins \\
Answer: A

\vspace{0.6em}
\noindent
Question: In 2008, Martin Winter founded which significant research center at the University of Münster? \\
Choices: A. Paul Scherrer Institute, B. Institute of Chemical Technology of Inorganic Materials, \\
C. MEET Battery Research Center, D. Institute of Physical Chemistry \\
Answer: C

\vspace{0.6em}
\noindent
Question: What did William Lyon Mackenzie King famously say about Canada's geography and history? \\
Choices: A. Canada has too much diversity, B. Canada has too much culture, \\
C. Canada has too much history, D. Canada has too much geography \\
Answer: D

\vspace{0.6em}
\noindent
Question: Who is the head chef of Gordon Ramsay au Trianon? \\
Choices: A. Frederic Larquemin, B. Joël Robuchon, C. Gordon Ramsay, D. Alain Ducasse \\
Answer: A

\vspace{0.6em}
\noindent
Question: What is the architectural style of St. John in the Wilderness in Nainital? \\
Choices: A. Modern, B. Gothic, C. Baroque, D. Renaissance \\
Answer: B

\vspace{0.8em}
\noindent
Question: \texttt{\{question\}} \\
Choices: \texttt{\{choice\_text\}}

\vspace{0.8em}
\noindent
Answer with only one letter (A, B, C, or D).
\end{tcolorbox}

\begin{tcolorbox}[
  colback=gray!5,
  colframe=gray!60,
  title= CoT Zero-shot Prompt,
  fonttitle=\bfseries,
  enhanced,
  breakable,
  boxrule=0.8pt,
  arc=3pt,
  center,
  before skip=4pt,
  after skip=4pt
]
\normalsize
\noindent
\textbf{System:} 

\vspace{0.4em}
You are given a multiple-choice question (A, B, C, D). \\
First, reason step by step and explain your thinking clearly. \\
Then, provide your final answer as only one letter: A, B, C, or D. \\ 
Always write the final answer in the format '\#\#\# FINAL ANSWER: X', where X is one of A, B, C, or D. \\
Do not include anything else after this line.

\vspace{1em}
\noindent
\textbf{User:}

\vspace{0.4em}
Question: \texttt{\{question\}} \\
Choices: \texttt{\{choice\_text\}}

\vspace{0.8em}
\noindent
Write out your reasoning step by step, and then write the final answer in this format:

\vspace{0.4em}
\noindent
\#\#\# FINAL ANSWER: 
\end{tcolorbox}

\begin{tcolorbox}[
  colback=gray!5,
  colframe=gray!60,
  title= CoT Few-shot Prompt (5-shot),
  fonttitle=\bfseries,
  enhanced,
  breakable,
  boxrule=0.8pt,
  arc=3pt,
  center,
  before skip=4pt,
  after skip=4pt
]
\normalsize
\noindent
\textbf{System:}

\vspace{0.4em}
You are given a multiple-choice question (A, B, C, D). \\
First, reason step by step and explain your thinking clearly. \\
Then, provide your final answer as only one letter: A, B, C, or D. \\ 
Always write the final answer in the format '\#\#\# FINAL ANSWER: X', where X is one of A, B, C, or D. \\
Do not include anything else after this line.

\vspace{1em}
\noindent
\textbf{User:}

\vspace{0.4em}
Question: Which geographical feature is located to the east and northeast of Scottsdale, Arizona? \\

\vspace{0.1em}
Choices: A. The Grand Canyon, B. The McDowell Mountain Range, C. The Sierra Madre Mountains, D. The Rocky Mountains \\

\vspace{0.1em}
Reasoning: Scottsdale, Arizona is located in the Sonoran Desert. The Grand Canyon is to the northwest of Scottsdale. The Sierra Madre Mountains are located in Mexico, far to the south. The Rocky Mountains are much further to the northeast, in states like Colorado and Wyoming. The McDowell Mountain Range is a series of mountains located to the northeast of Scottsdale, Arizona. This range is a prominent geographical feature in the area and is known for its hiking trails and scenic views. Therefore, the correct answer is B. The McDowell Mountain Range.\\
\#\#\# FINAL ANSWER: B \\
                    
\vspace{0.6em}
Question: \texttt{\{question\}} \\
Choices: \texttt{\{choice\_text\}}

\end{tcolorbox}

\subsection{Reasoning Domain Prompt}
\label{prompts:reasoning}


\begin{tcolorbox}[
  colback=gray!5,
  colframe=gray!60,
  title=Reasoning Prompt for Factual Domain,
  fonttitle=\bfseries,
  enhanced,
  breakable,
  boxrule=0.8pt,
  arc=3pt
]
\normalsize
\noindent
\textbf{System:} 

\vspace{0.4em}
You are a helpful assistant that solves math word problems. \\
You must reason step by step and produce a correct numerical answer.

\vspace{0.8em}
\noindent
\textbf{User:} 

\vspace{0.4em}
Solve the following math problem. \\
Question: \texttt{\{question\}} \\
Write out your reasoning step by step in plain text. \\
Then, on a new line, write the final answer in the following format:

\vspace{0.4em}
\noindent
\#\#\# FINAL ANSWER: $<$\texttt{number}$>$
\end{tcolorbox}



\section{Additional Experiments}

\subsection{Results in Code Dataset}

\subsubsection{Dataset}\label{sec:Results_in_Code_Dataset}
To evaluate code generation performance, we use two benchmarks 
from the unseen split: HumanEval~\citep{chen2021evaluating} 
(164 problems) and MBPP~\citep{austin2021program} (500 problems). 
Both datasets are treated as out-of-distribution test sets, 
as neither was included in the OLMo 2 pretraining corpus.


\subsubsection{Generalization in Code Dataset}\label{sec:code_dataset}

\textbf{RLVR Does Not Consistently Outperform SFT in Code Generation.}
The comprehensive evaluation results for code generation are presented in Table~\ref{tab:RQ1_code_datasetg}. Contrary to the consistent performance gains observed in math-based reasoning, RLVR exhibits limited advantages in code benchmarks.
On HumanEval, while RLVR and SFT achieve comparable Pass@1 performance, SFT's lead becomes more evident as $k$‬ increases.
This pattern is even more pronounced in MBPP, where SFT consistently surpasses RLVR across most settings, particularly in higher- $k$‬ regimes.
Overall, these findings suggest that RLVR provides limited benefits in low- $k$‬ exploitation settings and lacks a clear advantage over SFT in high-$k$‬ exploration for code generation.

\textbf{RLVR’s performance gains depend on reward--task alignment.}
The observed performance disparity highlights the critical importance of reward--task alignment.
While a significant RLVR--SFT gap exists in math benchmarks, this trend does not hold for code, where SFT may even outperform RLVR.
This suggests that RLVR's efficacy is most pronounced when the reward signal aligns closely with well-structured reasoning processes.
In contrast, for execution-dependent domains like code, these gains are diminished because correctness is less directly captured by the immediate reward signal.
Overall, RLVR is best understood not as a universal performance booster, but as a post-training method tailored for reinforcing reward-aligned reasoning.

\begin{table}[!htbp]
  \centering
  \caption{Code Dataset evaluation}
  \label{tab:RQ1_code_datasetg}
  \setlength{\tabcolsep}{5pt}
  \renewcommand{\arraystretch}{1.1}
  \footnotesize
  \resizebox{\columnwidth}{!}{%
    \begin{subtable}[t]{0.48\textwidth}
\centering
\caption{SFT}
\begin{tabular}{llcc}
\toprule
\textbf{Model} & \textbf{Metric} & \textbf{HumanEval} & \textbf{MBPP} \\
\midrule
\multirow{4}{*}{OLMo2 7B-SFT}
& Pass@1 & 41.96 & 28.29 \\
& Pass@2 & 45.31 & 29.96 \\
& Pass@4 & 47.95 & 31.36 \\
& Pass@8 & 50.58 & 32.60 \\
\midrule
\multirow{4}{*}{OLMo2 13B-SFT}
& Pass@1 & 44.78 & 35.27 \\
& Pass@2 & 50.01 & 37.68 \\
& Pass@4 & 54.94 & 39.52 \\
& Pass@8 & 59.00 & 41.09 \\
\midrule
\multirow{4}{*}{OLMo2 32B-SFT}
& Pass@1 & 53.43 & 39.83 \\
& Pass@2 & 58.11 & 42.09 \\
& Pass@4 & 62.23 & 43.98 \\
& Pass@8 & 65.69 & 46.07 \\
\bottomrule
\end{tabular}
\end{subtable}
\hfill
\begin{subtable}[t]{0.48\textwidth}
\centering
\caption{RLVR}
\begin{tabular}{llcc}
\toprule
\textbf{Model} & \textbf{Metric} & \textbf{HumanEval} & \textbf{MBPP} \\
\midrule
\multirow{4}{*}{OLMo2 7B}
& Pass@1 & 40.63 & 27.12 \\
& Pass@2 & 44.99 & 28.37 \\
& Pass@4 & 48.76 & 29.40 \\
& Pass@8 & 51.86 & 30.46 \\
\midrule
\multirow{4}{*}{OLMo2 13B}
& Pass@1 & 45.34 & 37.36 \\
& Pass@2 & 48.78 & 39.18 \\
& Pass@4 & 51.77 & 40.69 \\
& Pass@8 & 54.35 & 41.84 \\
\midrule
\multirow{4}{*}{OLMo2 32B}
& Pass@1 & 54.26 & 38.05 \\
& Pass@2 & 58.19 & 39.86 \\
& Pass@4 & 61.64 & 41.02 \\
& Pass@8 & 64.54 & 41.97 \\
\bottomrule
\end{tabular}
\end{subtable}
  }
\end{table}
\newpage
\subsection{Results under Chain-of-Thought Prompting} \label{sec:cot_head_tail_new}

\begin{table*}[h]
  \centering
  \caption{Head vs.\ Tail evaluation (with Gap) under Chain-of-Thought prompting (zero-shot and few-shot) (\%).} 
  \label{tab:RQ2_head_tail_gap2}

  \begingroup
  \setlength{\tabcolsep}{3pt}
  \renewcommand{\arraystretch}{0.92}
  \scriptsize

    \setlength{\tabcolsep}{10pt}
  \resizebox{\textwidth}{!}{%
    \setlength{\tabcolsep}{15pt}
\renewcommand{\arraystretch}{0.95}
{\small
\begin{tabular}{
  >{\centering\arraybackslash}m{1.2cm}   
  >{\centering\arraybackslash}m{2.2cm}   
  l                                     
  c c >{\columncolor{gray!8}}c          
  c c >{\columncolor{gray!8}}c          
}
\toprule
& & & \multicolumn{3}{c}{\textbf{CoT Zero-shot}}
    & \multicolumn{3}{c}{\textbf{CoT Few-shot (5)}} \\
\cmidrule(lr){4-6}\cmidrule(lr){7-9}

& & &
\textbf{Head} & \textbf{Tail} & \textbf{Gap}
& \textbf{Head} & \textbf{Tail} & \textbf{Gap} \\
\cmidrule(lr){4-4}\cmidrule(lr){5-5}\cmidrule(lr){6-6}
\cmidrule(lr){7-7}\cmidrule(lr){8-8}\cmidrule(lr){9-9}

\textbf{OLMo~2} & \textbf{Types} & \textbf{Metrics}
& & & 
& & & \\
\midrule

\multirow{7}{*}{\makecell{\textbf{7B}}}
& \multirow{3}{*}{\makecell{\textbf{Behavioral}}}
& Accuracy        & 82.40 & 50.29 & 32.12 & 87.94 & 60.27 & 27.67 \\
&  & $DC$     & 90.32 & 70.12 &  20.20 & 94.09 & 77.76 & 16.33 \\
&  & Self-Verify  & 28.54 & 45.03 & -16.49 & 24.38 & 35.51 &  -11.13 \\
\cmidrule{2-3}
& \multirow{4}{*}{\makecell{\textbf{Internal}\\$(PC)$}}
& Norm. $P$ & 81.31 & 65.74 & 15.57 & 76.45 & 62.29 & 14.16 \\
&  & MARS            & 61.35 & 65.22 & -3.87 & 61.59 & 64.92 & -3.33 \\
&  & Linear Probing  & 84.61 & 78.21 & 6.40 & 95.35 & 78.39 & 16.96 \\
\addlinespace[2pt]
\midrule

\multirow{7}{*}{\makecell{\textbf{13B}}}
& \multirow{3}{*}{\makecell{\textbf{Behavioral}}}
& Accuracy        & 88.67 & 62.58 & 26.09 & 87.53 & 60.69 & 26.84 \\
&  & $DC$     & 93.98 & 78.98 & 15.01 & 94.61 & 80.83 & 13.78 \\
&  & Self-Verify  & 42.52 & 38.95 & 3.57 & 66.71 & 55.54 & 11.17 \\
\cmidrule{2-3}
& \multirow{4}{*}{\makecell{\textbf{Internal}\\$(PC)$}}
& Norm. $P$ & 77.63 & 69.18 & 8.45 & 86.44 & 80.48 &  5.97 \\
&  & MARS            & 60.93 & 64.11 & -3.18 & 61.12 & 63.80 & -2.68 \\
&  & Linear Probing  & 93.69 & 82.59 & 11.10 & 83.83 & 82.35 & 1.48 \\
\addlinespace[2pt]
\midrule

\multirow{7}{*}{\makecell{\textbf{32B}}}
& \multirow{3}{*}{\makecell{\textbf{Behavioral}}}
& Accuracy        & 91.87 & 65.98 & 25.90 & 91.44 & 67.15 & 24.29 \\
&  & $DC$     & 97.02 & 84.78 & 12.24 & 97.75 & 84.15 & 13.61 \\
&  & Self-Verify  & 91.54 & 66.70 & 24.85 & 91.59 & 68.03 & 23.57 \\
\cmidrule{2-3}
& \multirow{4}{*}{\makecell{\textbf{Internal}\\$(PC)$}}
& Norm. $P$ & 73.85 & 60.69 & 13.15 & 75.41 & 64.63 & 10.78 \\
&  & MARS            & 59.12 & 65.88 & -6.76 & 56.52 & 65.53 & -9.01 \\
&  & Linear Probing  & 96.79 & 82.98 & 13.81 & 96.23 & 84.92 & 11.31 \\
\bottomrule
\end{tabular}}

  }

  \endgroup
\end{table*}

\newpage
\subsection{Results on the Cutoff-Only Unseen Set}
\label{sec:cutoff_only}

\textcolor{black}{
Unlike the failure-filtered subset used in the main analysis, the \emph{cutoff-only} unseen set retains questions independently of model performance, avoiding the selection bias introduced by failure-based filtering. However, some questions may still be answerable using pre-cutoff knowledge.
Overall, several key qualitative patterns from the main analysis remain consistent, including the seen--unseen gaps in Accuracy and Decisional Consistency, as well as the self-verification gap at 32B. Nevertheless, the magnitude of these gaps and some internal metrics differ from the filtered results.
}

\begin{table*}[h]
  \centering
  \caption{Factual evaluation across Seen/Unseen splits under standard prompting. Self-verify denotes the accuracy computed based on the corresponding confusion matrix.} 

  \begingroup
  \setlength{\tabcolsep}{3pt}
  \renewcommand{\arraystretch}{1.15}
  \small

  \resizebox{\textwidth}{!}{%
    \setlength{\tabcolsep}{6pt}
\renewcommand{\arraystretch}{1.15}
\resizebox{\textwidth}{!}{
\begin{tabular}{
  >{\centering\arraybackslash}m{1.6cm}
  >{\centering\arraybackslash}m{2.2cm}
  l
  c c >{\columncolor{gray!8}}c >{\columncolor{gray!8}}c
  c c >{\columncolor{gray!8}}c >{\columncolor{gray!8}}c
}

\toprule
& & & \multicolumn{4}{c}{\textbf{Zero-shot}}
    & \multicolumn{4}{c}{\textbf{Few-shot (5)}} \\
\cmidrule(lr){4-7} \cmidrule(lr){8-11}

& & &
\multicolumn{2}{c}{Seen} &
\multicolumn{2}{c}{Unseen} &
\multicolumn{2}{c}{Seen} &
\multicolumn{2}{c}{Unseen} \\
\cmidrule(lr){4-5}
\cmidrule(lr){6-7}
\cmidrule(lr){8-9}
\cmidrule(lr){10-11}

\textbf{\textbf{OLMo~2}} & \textbf{Types} & \textbf{Metrics}
& \makecell{\textbf{Academic}}
& \makecell{\textbf{Web}}
& \makecell{\textbf{RealTime}\\\textbf{QA}}
& \makecell{\textbf{Current}\\\textbf{events}}
& \makecell{\textbf{Academic}}
& \makecell{\textbf{Web}}
& \makecell{\textbf{RealTime}\\\textbf{QA}}
& \makecell{\textbf{Current}\\\textbf{events}} \\
\midrule

\multirow{8}{*}{\makecell{\textbf{7B}}}
& \multirow{3}{*}{\makecell{\textbf{Behavioral}}}
& Accuracy        & 75.53 & 70.09 & 24.59 & 47.51 & 78.99 & 76.04 & 26.31 & 52.35 \\
&  & $DC$     & 86.75 & 83.36 & 67.86 & 77.30 & 91.47 & 88.96 & 72.47 & 83.58 \\
&  & Self-Verify  & 28.00 & 34.29 & 61.27 & 58.86 & 26.52 & 29.77 & 69.58 & 56.34 \\
\cmidrule{2-3}
& \multirow{4}{*}{\makecell{\textbf{Internal}\\($PC$)}}
& Norm. $P$ & 63.19 & 57.49 & 8.48 & 26.77 & 63.44 & 56.82 & 7.77 & 26.46 \\
&  & MARS            & 58.73 & 60.00 & 58.37 & 48.27 & 58.45 & 62.20 & 58.59 & 47.16 \\
&  & Linear Probing  & 80.23 & 83.31 & 41.58 & 80.84 & 75.42 & 81.50 & 43.23 & 78.05 \\
&  & SAPLMA          & 75.72 & 83.24 & 37.63 & 67.07 & - & - & - & - \\
\addlinespace[2pt]
\midrule

\multirow{8}{*}{\makecell{\textbf{13B}}}
& \multirow{3}{*}{\makecell{\textbf{Behavioral}}}
& Accuracy        & 79.78 & 74.33 & 31.98 & 52.56 & 79.02 & 75.06 & 32.65 & 52.01 \\
&  & $DC$     & 91.07 & 85.19 & 55.64 & 77.97 & 89.89 & 86.36 & 57.16 & 78.52 \\
&  & Self-Verify  & 62.54 & 55.70 & 57.24 & 55.08 & 60.89 & 57.18 & 58.14 & 55.70 \\
\cmidrule{2-3}
& \multirow{4}{*}{\makecell{\textbf{Internal}\\($PC$)}}
& Norm. $P$ & 29.69 & 29.54 & 11.68 & 16.73 & 29.60 & 29.36 & 11.04 & 16.33 \\
&  & MARS            & 64.84 & 63.71 & 61.80 & 51.15 & 64.17 & 65.97 & 62.58 & 52.34 \\
&  & Linear Probing  & 76.15 & 82.53 & 55.70 & 82.67 & 80.89 & 83.55 & 49.06 & 86.21 \\
&  & SAPLMA          & 82.97 & 84.40 & 43.44 & 74.81 & - & - & - & - \\
\addlinespace[2pt]
\midrule

\multirow{8}{*}{\makecell{\textbf{32B}}}
& \multirow{3}{*}{\makecell{\textbf{Behavioral}}}
& Accuracy        & 82.97 & 80.22 & 41.78 & 60.52 & 82.41 & 79.92 & 38.12 & 58.44 \\
&  & $DC$     & 94.73 & 89.72 & 64.86 & 85.40 & 92.56 & 87.90 & 61.29 & 81.27 \\
&  & Self-Verify  & 83.76 & 79.67 & 66.88 & 64.30 & 83.83 & 80.53 & 66.84 & 65.04 \\
\cmidrule{2-3}
& \multirow{4}{*}{\makecell{\textbf{Internal}\\($PC$)}}
& Norm. $P$ & 57.06 & 54.15 & 14.07 & 34.79 & 57.52 & 54.73 & 14.01 & 35.95 \\
&  & MARS            & 56.91 & 62.45 & 62.24 & 50.40 & 56.00 & 62.86 & 64.05 & 51.99 \\
&  & Linear Probing  & 71.32 & 78.29 & 51.22 & 82.99 & 75.41 & 85.31 & 54.32 & 85.87 \\
&  & SAPLMA          & 82.61 & 86.19 & 45.59 & 74.96 & - & - & - & - \\
\bottomrule

\end{tabular}}
  }
  
  \endgroup
\end{table*}


\begin{table*}[h]
  \centering
  \caption{Self-Verify analysis results under standard prompting.}

  \begingroup
  \setlength{\tabcolsep}{3pt}
  \renewcommand{\arraystretch}{0.92}
  \scriptsize

    \setlength{\tabcolsep}{10pt}
  \resizebox{\textwidth}{!}{%
    \setlength{\tabcolsep}{6pt}
\renewcommand{\arraystretch}{0.8}
\resizebox{\textwidth}{!}{
\begin{tabular}{
  >{\centering\arraybackslash}m{1.1cm}
  l
  c c >{\columncolor{gray!8}}c >{\columncolor{gray!8}}c
  c c >{\columncolor{gray!8}}c >{\columncolor{gray!8}}c
}

\toprule
& & \multicolumn{4}{c}{\textbf{Zero-shot}}
  & \multicolumn{4}{c}{\textbf{Few-shot (5)}} \\
\cmidrule(lr){3-6} \cmidrule(lr){7-10}

& &
\multicolumn{2}{c}{Seen} &
\multicolumn{2}{c}{Unseen} &
\multicolumn{2}{c}{Seen} &
\multicolumn{2}{c}{Unseen} \\
\cmidrule(lr){3-4}
\cmidrule(lr){5-6}
\cmidrule(lr){7-8}
\cmidrule(lr){9-10}

\textbf{OLMo~2} & \textbf{Metrics}
& \makecell{\textbf{Academic}}
& \makecell{\textbf{Web}}
& \makecell{\textbf{RealTime}\\\textbf{QA}}
& \makecell{\textbf{Current}\\\textbf{events}}
& \makecell{\textbf{Academic}}
& \makecell{\textbf{Web}}
& \makecell{\textbf{RealTime}\\\textbf{QA}}
& \makecell{\textbf{Current}\\\textbf{events}} \\
\midrule

\multirow[c]{4}{*}{\makecell[c]{\textbf{7B}}}
& Accuracy        & 28.00 & 34.29 & 70.77 & 58.86 & 26.52 & 29.77 & 69.58 & 56.34 \\
& Precision            & 14.13 & 14.62 & 22.68 & 23.33 & 12.68 & 13.30 & 21.91 & 21.40 \\
& Recall     & 14.13 & 14.62 & 28.39 & 23.37 & 12.68 & 13.33 & 27.53 & 21.40 \\
& F1      & 14.13 & 14.62 & 23.96 & 23.35 & 12.68 & 13.31 & 23.16 & 21.40 \\

\midrule

\multirow[c]{4}{*}{\makecell[c]{\textbf{13B}}}
& Accuracy        & 62.54 & 55.70 & 57.24 & 55.08 & 60.89 & 57.18 & 58.14 & 55.70 \\
& Precision            & 56.64 & 50.32 & 37.10 & 64.30 & 55.90 & 51.31 & 66.84 & 65.04 \\
& Recall     & 61.45 & 57.38 & 57.63 & 57.51 & 61.81 & 56.81 & 55.70 & 57.36 \\
& F1       & 57.89 & 52.09 & 41.83 & 48.19 & 57.45 & 52.79 & 41.23 & 47.30 \\

\midrule

\multirow[c]{4}{*}{\makecell[c]{\textbf{32B}}}
& Accuracy        & 83.76 & 79.67 & 66.88 & 64.30 & 83.83 & 80.53 & 66.84 & 65.04 \\
& Precision        & 79.69 & 76.12 & 37.02 & 58.10 & 79.55 & 76.58 & 33.79 & 57.21 \\
& Recall     & 81.88 & 82.32 & 41.29 & 67.37 & 82.25 & 83.18 & 38.71 & 68.59 \\
& F1       & 80.22 & 77.80 & 38.13 & 60.46 & 80.34 & 78.45 & 34.99 & 59.92 \\

\bottomrule

\end{tabular}
}

  }
  \endgroup
\end{table*}


\begin{table*}[h]
  \centering
  \caption{Factual evaluation across Seen/Unseen splits under Chain-of-Thought prompting.}

  \begingroup
  \setlength{\tabcolsep}{3pt}
  \renewcommand{\arraystretch}{1.15}
  \small

  \resizebox{\textwidth}{!}{%
    \setlength{\tabcolsep}{6pt}
\renewcommand{\arraystretch}{1.15}
\resizebox{\textwidth}{!}{
\begin{tabular}{
  >{\centering\arraybackslash}m{1.6cm}
  >{\centering\arraybackslash}m{2.2cm}
  l
  c c >{\columncolor{gray!8}}c >{\columncolor{gray!8}}c
  c c >{\columncolor{gray!8}}c >{\columncolor{gray!8}}c
}

\toprule
& & & \multicolumn{4}{c}{\textbf{CoT Zero-shot}}
    & \multicolumn{4}{c}{\textbf{CoT Few-shot (5)}} \\
\cmidrule(lr){4-7} \cmidrule(lr){8-11}

& & &
\multicolumn{2}{c}{Seen} &
\multicolumn{2}{c}{Unseen} &
\multicolumn{2}{c}{Seen} &
\multicolumn{2}{c}{Unseen} \\
\cmidrule(lr){4-5}
\cmidrule(lr){6-7}
\cmidrule(lr){8-9}
\cmidrule(lr){10-11}

\textbf{\textbf{OLMo~2}} & \textbf{Types} & \textbf{Metrics}
& \makecell{\textbf{Academic}}
& \makecell{\textbf{Web}}
& \makecell{\textbf{RealTime}\\\textbf{QA}}
& \makecell{\textbf{Current}\\\textbf{events}}
& \makecell{\textbf{Academic}}
& \makecell{\textbf{Web}}
& \makecell{\textbf{RealTime}\\\textbf{QA}}
& \makecell{\textbf{Current}\\\textbf{events}} \\
\midrule

\multirow{8}{*}{\makecell{\textbf{7B}}}
& \multirow{3}{*}{\makecell{\textbf{Behavioral}}}
& Accuracy        & 71.25 & 66.34 & 23.21 & 44.67 & 75.63 & 74.10 & 24.73 & 53.21 \\
&  & $DC$     & 83.09 & 80.22 & 62.07 & 71.89 & 89.47 & 85.92 & 67.51 & 79.61 \\
&  & Self-Verify  & 31.95 & 36.79 & 64.34 & 60.39 & 26.09 & 29.94 & 64.73 & 53.18 \\
\cmidrule{2-3}
& \multirow{4}{*}{\makecell{\textbf{Internal}\\($PC$)}}
& Norm. $P$ & 73.69 & 73.53 & 13.67 & 50.63 & 74.82 & 69.37 & 8.92 & 56.66 \\
&  & MARS            & 59.84 & 63.29 & 59.16 & 49.37 & 59.00 & 63.26 & 56.31 & 48.49 \\
&  & Linear Probing  & 80.53 & 86.45 & 48.96 & 79.23 & 75.37 & 79.58 & 37.36 & 76.95 \\
\addlinespace[2pt]
\midrule

\multirow{8}{*}{\makecell{\textbf{13B}}}
& \multirow{3}{*}{\makecell{\textbf{Behavioral}}}
& Accuracy        & 79.78 & 75.63 & 33.74 & 57.03 & 78.69 & 74.11 & 34.06 & 59.21 \\
&  & $DC$     & 89.41 & 86.48 & 64.74 & 79.96 & 91.32 & 87.72 & 64.93 & 82.77 \\
&  & Self-Verify  & 38.20 & 40.73 & 67.27 & 54.19 & 60.14 & 61.13 & 62.31 & 61.84 \\
\cmidrule{2-3}
& \multirow{4}{*}{\makecell{\textbf{Internal}\\($PC$)}}
& Norm. $P$ & 70.68 & 73.40 & 30.97 & 62.94 & 81.01 & 83.46 & 41.08 & 76.99 \\
&  & MARS            & 59.41 & 62.52 & 64.43 & 49.94 & 57.24 & 62.46 & 60.42 & 50.60 \\
&  & Linear Probing  & 78.39 & 85.04 & 52.65 & 84.00 & 79.71 & 71.71 & 51.95 & 79.81 \\
\addlinespace[2pt]
\midrule

\multirow{8}{*}{\makecell{\textbf{32B}}}
& \multirow{3}{*}{\makecell{\textbf{Behavioral}}}
& Accuracy        & 82.41 & 78.93 & 40.88 & 61.02 & 82.18 & 79.30 & 44.95 & 62.23 \\
&  & $DC$     & 94.41 & 90.90 & 69.23 & 87.07 & 93.97 & 90.95 & 72.81 & 86.96 \\
&  & Self-Verify  & 83.56 & 79.12 & 64.46 & 64.11 & 82.28 & 79.81 & 63.36 & 65.03 \\
\cmidrule{2-3}
& \multirow{4}{*}{\makecell{\textbf{Internal}\\($PC$)}}
& Norm. $P$ & 68.14 & 67.27 & 21.74 & 48.99 & 69.80 & 70.02 & 16.96 & 54.99 \\
&  & MARS            & 56.19 & 62.50 & 60.20 & 50.16 & 54.84 & 61.03 & 58.70 & 50.33 \\
&  & Linear Probing  & 68.78 & 77.80 & 59.79 & 82.02 & 72.55 & 79.88 & 54.39 & 81.31 \\

\bottomrule
\end{tabular}}

  }
  
  \endgroup
\end{table*}

\begin{table*}[h]
  \centering
  \caption{Self-Verify analysis results under Chain-of-Thought prompting}

  \begingroup
  \setlength{\tabcolsep}{3pt}
  \renewcommand{\arraystretch}{0.92}
  \scriptsize

    \setlength{\tabcolsep}{10pt}
  \resizebox{\textwidth}{!}{%
    \setlength{\tabcolsep}{6pt}
\renewcommand{\arraystretch}{0.8}
\resizebox{\textwidth}{!}{
\begin{tabular}{
  >{\centering\arraybackslash}m{1.1cm}
  l
  c c >{\columncolor{gray!8}}c >{\columncolor{gray!8}}c
  c c >{\columncolor{gray!8}}c >{\columncolor{gray!8}}c
}

\toprule
& & \multicolumn{4}{c}{\textbf{CoT Zero-shot}}
  & \multicolumn{4}{c}{\textbf{CoT Few-shot (5)}} \\
\cmidrule(lr){3-6} \cmidrule(lr){7-10}

& &
\multicolumn{2}{c}{Seen} &
\multicolumn{2}{c}{Unseen} &
\multicolumn{2}{c}{Seen} &
\multicolumn{2}{c}{Unseen} \\
\cmidrule(lr){3-4}
\cmidrule(lr){5-6}
\cmidrule(lr){7-8}
\cmidrule(lr){9-10}

\textbf{OLMo~2} & \textbf{Metrics}
& \makecell{\textbf{Academic}}
& \makecell{\textbf{Web}}
& \makecell{\textbf{RealTime}\\\textbf{QA}}
& \makecell{\textbf{Current}\\\textbf{events}}
& \makecell{\textbf{Academic}}
& \makecell{\textbf{Web}}
& \makecell{\textbf{RealTime}\\\textbf{QA}}
& \makecell{\textbf{Current}\\\textbf{events}} \\
\midrule

\multirow{4}{*}{\makecell{\textbf{7B}}}
& Accuracy        & 31.95 & 36.79 & 64.34 & 60.39 & 26.09 & 29.94 & 64.73 & 53.18 \\
& Precision            & 12.70 & 15.69 & 29.65 & 26.25 & 18.57 & 22.02 & 24.06 & 32.51 \\
& Recall     & 12.46 & 14.69 & 38.14 & 26.40 & 13.33 & 16.22 & 23.73 & 28.95 \\
& F1      & 12.47 & 14.70 & 29.88 & 26.04 & 14.13 & 16.95 & 21.37 & 29.72 \\

\midrule

\multirow{4}{*}{\makecell{\textbf{13B}}}
& Accuracy        & 38.20 & 40.73 & 67.27 & 54.19 & 60.14 & 61.13 & 62.31 & 61.84 \\
& Precision            & 52.33 & 53.01 & 42.97 & 53.03 & 73.23 & 70.82 & 43.39 & 60.55 \\
& Recall     & 32.68 & 38.54 & 49.12 & 51.57 & 62.63 & 69.63 & 55.94 & 76.36 \\
& F1       & 35.66 & 39.39 & 43.35 & 47.10 & 63.01 & 65.11 & 45.47 & 63.38 \\

\midrule

\multirow{4}{*}{\makecell{\textbf{32B}}}
& Accuracy        & 83.56 & 79.12 & 64.46 & 64.11 & 82.28 & 79.81 & 63.36 & 65.03 \\
& Precision        & 82.62 & 79.01 & 47.39 & 61.86 & 82.73 & 79.63 & 47.78 & 62.61 \\
& Recall     & 86.93 & 86.82 & 53.69 & 72.05 & 85.14 & 85.63 & 51.36 & 72.43 \\
& F1       & 83.37 & 80.29 & 47.68 & 63.43 & 82.44 & 80.17 & 46.97 & 63.95 \\

\bottomrule

\end{tabular}

}
  }

  \endgroup
\end{table*}



\end{document}